\documentclass{article}

    \PassOptionsToPackage{square, sort&compress, numbers}{natbib}

\usepackage[preprint]{neurips_2022}

\usepackage[utf8]{inputenc} %
\usepackage[T1]{fontenc}    %
\usepackage[]{hyperref}       %
\usepackage{url}            %
\usepackage{booktabs}       %
\usepackage{amsfonts}       %
\usepackage{nicefrac}       %
\usepackage{microtype}      %
\usepackage{xcolor}         %
\usepackage{multirow}
\usepackage{graphicx}
\usepackage{float}
\usepackage{caption}
\usepackage{subcaption}
\usepackage{cleveref}
\usepackage{algorithmic, algorithm}
\usepackage{enumitem}

\graphicspath{ {./figures/} }

\title{Group SELFIES: A Robust Fragment-Based Molecular String Representation}

\author{%
  Austin H. Cheng\thanks{Equal contribution.} \thanks{Correspondence to: \texttt{austin@cs.toronto.edu}} \\
  University of Toronto
   \And
   Andy Cai\footnotemark[1] \\
   University of Toronto \\
  \And
  Santiago Miret \\
    Intel Labs\\
      \And
    Gustavo Malkomes \\
    SigOpt\\
    \And
    Mariano Phielipp \\
    Intel Labs\\
      \And
    Alán Aspuru-Guzik \\
    University of Toronto\\\
}

\begin{document}

\maketitle

\begin{abstract}
    We introduce Group SELFIES, a molecular string representation that leverages group tokens to represent functional groups or entire substructures while maintaining chemical robustness guarantees. Molecular string representations, such as SMILES and SELFIES, serve as the basis for molecular generation and optimization in chemical language models, deep generative models, and evolutionary methods. While SMILES and SELFIES leverage atomic representations, Group SELFIES builds on top of the chemical robustness guarantees of SELFIES by enabling group tokens, thereby creating additional flexibility to the representation. Moreover, the group tokens in Group SELFIES can take advantage of inductive biases of molecular fragments that capture meaningful chemical motifs. The advantages of capturing chemical motifs and flexibility are demonstrated in our experiments, which show that Group SELFIES improves distribution learning of common molecular datasets. Further experiments also show that random sampling of Group SELFIES strings improves the quality of generated molecules compared to regular SELFIES strings. Our open-source implementation of Group SELFIES is available at \url{https://github.com/aspuru-guzik-group/group-selfies}, which we hope will aid future research in molecular generation and optimization.
\end{abstract}

\section{Introduction} \label{sec:intro}

The discovery of functional molecules for drugs and energy materials is crucial to tackling global challenges in public health and climate change. Different types of generative models can suggest potential molecules to synthesize and test, but the performance of the models and molecules heavily relies on the underlying molecular representation. Several models generate molecules represented as SMILES strings \citep{chithrananda2020chemberta, gomez2018automatic, blaschke2020reinvent, moss2020boss, sanchez2017optimizing}, but their generated output can be invalid due to syntax errors or incorrect valency. SELFIES \cite{krenn2020self} is a molecular string representation that overcomes chemical invalidity challenges by ensuring that any string of SELFIES characters can be decoded to a molecule with valid valency. This not only makes it a natural representation for chemical language models that output molecular strings, but also for genetic algorithms such as GA+D, STONED, and JANUS \citep{nigam2019augmenting,nigam2021beyond,nigam2022parallel} for molecular optimization.

SELFIES improves string-based molecular generation by encoding prior knowledge of valency constraints into the representation independently of the optimization method. The representation has been shown to improve distribution learning by language models \citep{flam2022language}, as well as image2string and string2string translation \cite{rajan2020decimer, rajan2021stout} and molecular generation in data-sparse regimes \citep{frey2022fastflows}. Additionally, simple add/replace/delete edits to SELFIES strings can generate new but similar molecules, enabling genetic algorithms that directly manipulate strings to generate molecules \cite{nigam2019augmenting}. Alternatively, guiding these simple string edits with Tanimoto similarity can interpolate between molecules as performed in STONED by \citet{nigam2021beyond}, which can then be applied as crossover operations in genetic algorithms such as JANUS \citep{nigam2022parallel}. Molecular interpolation has also been used to find counterfactual decision boundaries that explain a molecular classifier's decisions \citep{wellawatte2022model}.

While SMILES and SELFIES represent molecules at the individual atom and bond level, human chemists typically think about molecules in terms of the substructures that they contain. Human chemists can distinguish molecular substructures based solely on the image of a molecule and induce the molecular properties those substructures usually imply. Many fragment-based generative models take advantage of this inductive bias \cite{jensen2019graph, jin2018junction, xie2021mars, bengio2021flow, jin2020multi, yang2021hit, flam2022scalable, guo2022data, liu2018constrained, polishchuk2020crem} by constructing custom representations amenable to fragment-based molecular design. However, these approaches are not string-based, thereby losing desirable properties of string representations: easy manipulation, and direct input into established language models.

Similar to how SELFIES incorporates prior knowledge of valency constraints, we can also incorporate prior knowledge in the form of functional groups and molecular substructures into the representation. In this work, we combine the flexibility of string representations with the chemical robustness of SELFIES and the interpretability and inductive bias of fragment-based approaches into a novel string representation: Group SELFIES, a robust string representation that extends SELFIES to include tokens which represent functional groups or entire substructures.

\begin{figure}[t]
    \centering
    \includegraphics[width=\textwidth]{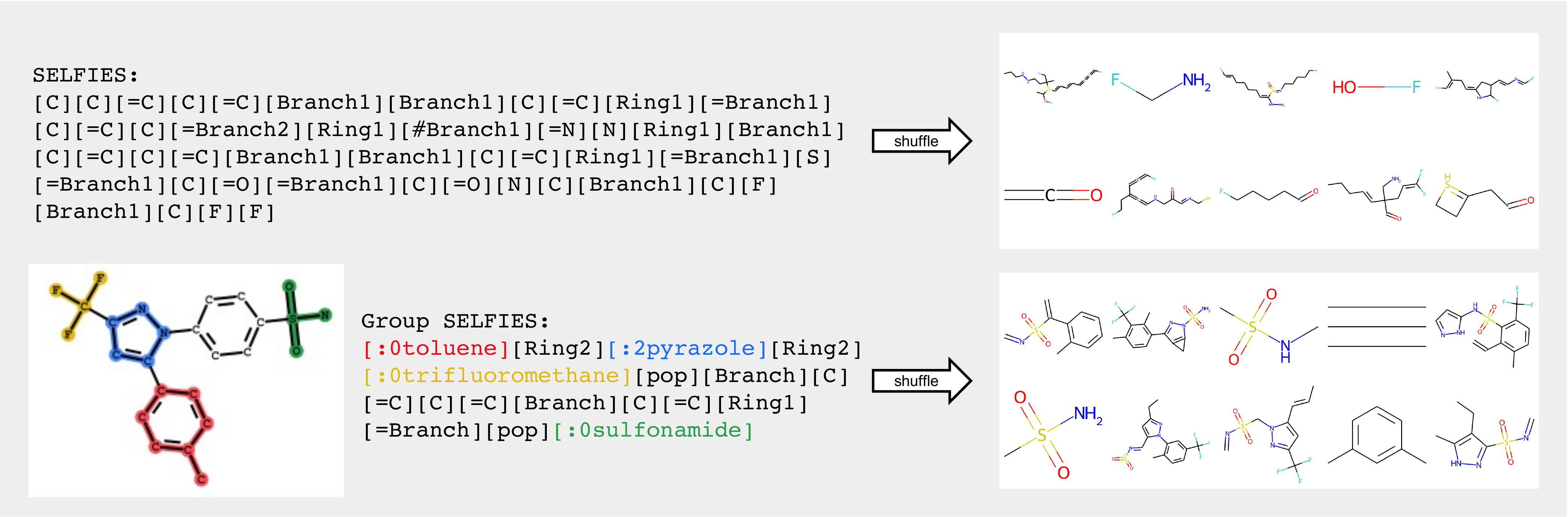}
    \caption{Visual overview of SELFIES and Group SELFIES. SELFIES is robust, so shuffling tokens around will yield new molecules with correct valency. Group SELFIES maintains robustness while adding group tokens, highlighted in color. When Group SELFIES tokens are shuffled, structures like benzene rings are more often preserved, while shuffled SELFIES strings rarely ever preserve structures. Incidentally, Group SELFIES also improves the readability of molecular string representations since chemists can see what substructures are present.}
    \label{fig:overview}
\end{figure}

In \Cref{sec:related-work}, we discuss how Group SELFIES fits into related research and then formally introduce the representation in \Cref{sec:representation}. Specifically, we outline how molecules are encoded into and decoded from Group SELFIES, and we show that arbitrary Group SELFIES strings can be decoded to molecules with valid valency. The representation enables users to easily specify their own groups or extract fragments from a dataset, leveraging the wide area of cheminformatics research available there. In \Cref{sec:experiments}, we find that Group SELFIES is more compact than SMILES or SELFIES and improves distribution learning. Additionally, we compare molecules generated via randomly sampling SELFIES and Group SELFIES strings and find that Group SELFIES improves the quality of generated molecules. Molecular generation via random sampling provides greater emphasis on the representation itself by abstracting away the complexities of the type of generative method used, which we leave to future work as described in \Cref{sec:discussion}.

Overall, Group SELFIES provides the flexibility of \emph{group representation}, the ability to represent \emph{extended chirality} via chiral group tokens and \emph{chemical robustness} as summarized in \Cref{comparison}.

\begin{table}[ht]
\centering
\begin{tabular}{ccccc}
\toprule
Representation & Robustness & \begin{tabular}{c}Substructure \\ Control\end{tabular} & \begin{tabular}{c}Extended \\ Chirality\end{tabular}  & \begin{tabular}{c}Distribution \\ Learning\end{tabular}\\
\midrule
SMILES & no & no & no & \textasciitilde \\
SELFIES & yes & no & no & \textasciitilde \\
Group SELFIES & yes & yes & yes & improved \\
\bottomrule \\
\end{tabular}

{\centering
\caption{ \label{comparison} Comparison of the capabilities of SMILES, SELFIES, and Group SELFIES. Group SELFIES provides \emph{group representation}, representation of \emph{extended chirality} and \emph{chemical robustness}. Additionally as shown in \Cref{sec:experiments}, Group SELFIES improves distribution learning compared to other representations.}
}
\end{table}

\section{Related Work} \label{sec:related-work}

\textbf{Fragment-Based String Representations:} Group SELFIES is not the first fragment-based string representation that has been proposed. Historical string representations, such as Wiswesser Line Notation (WLN) \cite{wiswesser1951simplified, wiswesser1968107, vollmer1983wiswesser}, Hayward Notation \cite{hayward1961new}, and Skolnik Notation \cite{skolnik1964notation}, all predate SMILES and represent molecules non-atomically. They use tokens that represent functional groups, such as carboxyls or phenyls, as well as ring systems. WLN strings are usually shorter and sometimes easier for trained humans to understand than SMILES, as it is easier to recognize functional groups encoded as single characters than functional groups encoded atomically. 
SYBYL Line Notation (SLN) \cite{homer2008sybyl} allows for ``macro atoms'' which specify multiple atoms in a substructure. The Hierarchical Editing Language for Macromolecules (HELM) \cite{zhang2012helm} represents complex biomolecules by declaring monomers and then connecting them in a polymer line notation. Human-Readable SMILES \cite{garay2021human} applies common abbreviations for chemical substituents to process and compress SMILES strings into a more human-readable format. SMILES Pair Encoding \cite{li2021smiles} breaks down SMILES strings by tokenizing them in a data-driven way that recognizes common substructures.

\textbf{Genetic Programming:} A string representation of molecules such as SELFIES can be thought of as a \textit{programming language} where programs specify how to construct molecules. \textit{Genetic programming} \cite{koza2005genetic} uses genetic algorithms to design programs that fulfill desired constraints. In particular for linear genetic programming \cite{brameier2007linear}, programs are represented as linear sequences of atomic instructions, rather than as a tree of expressions. Linear sequences of atomic instructions are easily mutated by changing any instruction in the sequence, since any sequence of atomic instructions will still be a runnable program. In this way, SELFIES and Group SELFIES can be thought of as domain-specific languages for linear genetic programming.

\textbf{Learned Grammars:} Data-Efficient Graph Grammar Learning (DEG) \cite{guo2022data} is a recent approach for extracting useful formal graph grammars from small datasets of molecules. In this context, a ``useful grammar'' means that molecules generated by applying random applicable production rules usually have high scores. The learned production rules of the graph grammar can be thought of as similar to functional groups applied in Group SELFIES. Group SELFIES allows for flexibility and fine control of substructures, which can extracted from any procedure including DEG.

\section{Representation} \label{sec:representation}

\subsection{SELFIES Framework}

\begin{figure}
    \centering
    \includegraphics[width=\textwidth]{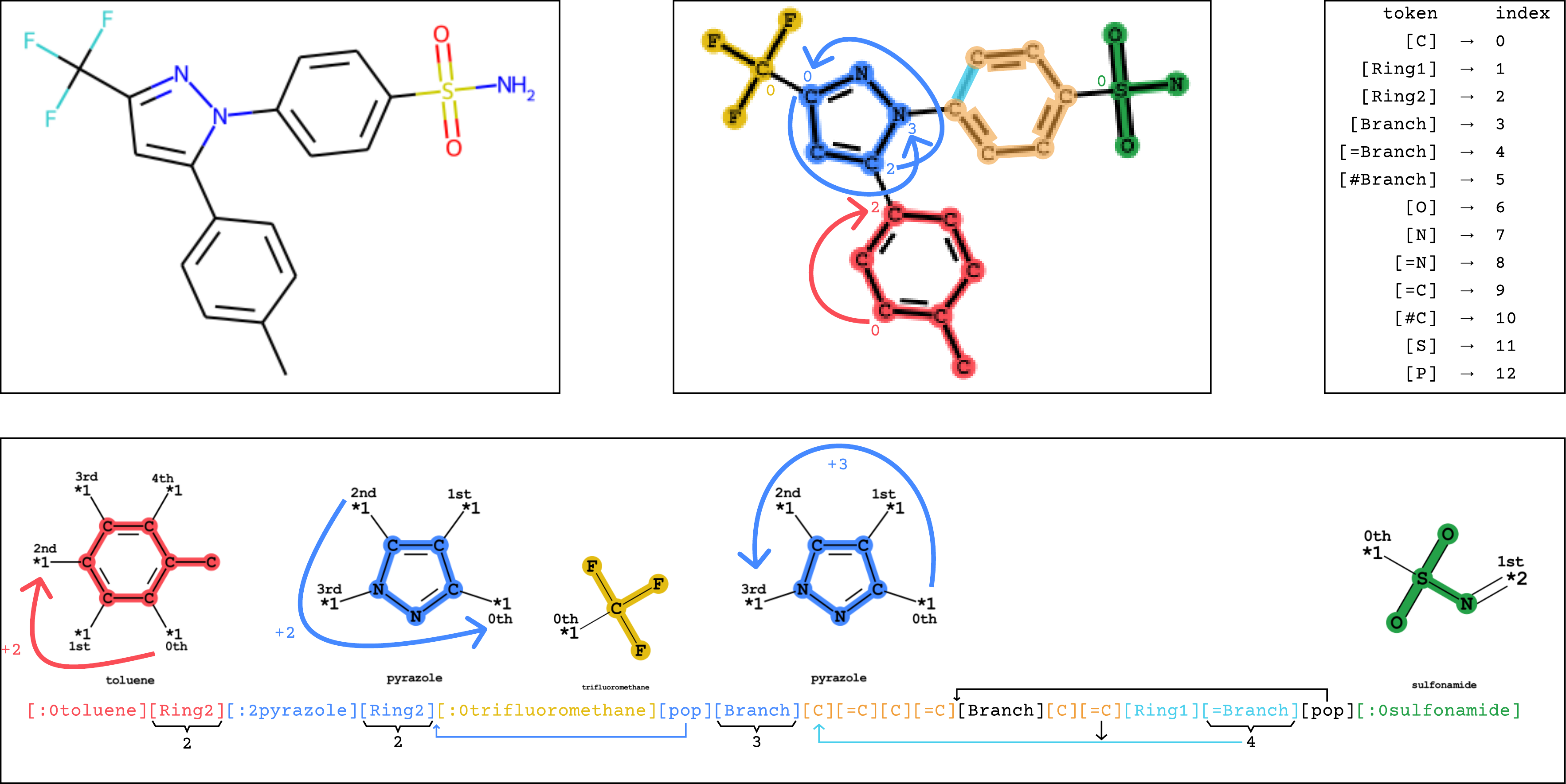}
    \caption{Visual explanation of Group SELFIES encoding/decoding of celecoxib. Top-left: molecular structure of celecoxib. Top-middle: the structure of celecoxib colored by its groups and atoms, with arrows and attachment indices indicating how the encoder and decoder navigate around the groups. Top-right: index overload table in Group SELFIES, indicating how tokens are interpreted as numbers. Bottom: Celecoxib represented in Group SELFIES. Tokens are colored by the groups and atoms they act on. Index overloads are shown where interpreted. Arrows indicate how branches return to their original branchpoints and how rings form bonds with previously placed atoms (light-blue).}
    \label{fig:explain}
\end{figure}

Before introducing in Group SELFIES in greater detail, we summarize the primary features of SELFIES and the reasons underlying its chemical robustness. SELFIES is equipped with an encoder and a decoder. The encoder takes in a molecule and converts it to a SELFIES string, and the decoder takes in a SELFIES string and converts it to a molecule. To encode a molecule in SELFIES, one traverses its molecular graph and outputs the processed traversal as a string of SELFIES tokens. To decode a SELFIES string, one reads through the string token-by-token, building the molecular graph along the way until arriving at the finished graph. Since the encoding and decoding process alone does not guarantee chemical robustness, the SELFIES decoder further includes two important features:

\begin{enumerate}
    \item Each token in SELFIES is overloaded to ensure that it can be interpreted sensibly in all contexts. For instance, all tokens in SELFIES can also be interpreted as numbers, which is useful when expressing branch and ring lengths.
    \item SELFIES keeps track of the available valency at each step in the decoding process; if a bond would be formed that would exceed this valency, it changes the bond order or ignores the bond. For instance, when decoding \texttt{[C][O][=C]}, adding \texttt{[=C]} would exceed the valency of \texttt{[O]}, so SELFIES changes the bond order and adds \texttt{[C]} instead. 
\end{enumerate}

By preserving these properties in the Group SELFIES decoder, we ensure chemical robustness is preserved.

\subsection{Basic Tokens in Group SELFIES}
Group SELFIES strings consist of the following fundamental tokens:
\begin{itemize}[leftmargin=1em]
    \item \texttt{[X]} adds an atom with the atomic symbol X.
    \item \texttt{[Branch]} creates a new branch off the current atom and saves the current atom as a branchpoint to return to later, and is analogous to an opening parenthesis \texttt{(} in SMILES. \texttt{[pop]} exits the current branch, returning to the most recent branchpoint, and is analogous to a closing parenthesis \texttt{)} in SMILES. Unlike in SMILES, however, \texttt{[Branch]} and \texttt{[pop]} tokens need not come in pairs, which helps maintain robustness. This token is also different from the \texttt{[BranchX]} tokens in SELFIES. Experiments in Appendix \ref{no_group} indicate this change does not substantially affect the performance of Group SELFIES.
    \item \texttt{[RingX]} indicates that a ring bond will be formed from the current atom. The next \texttt{X} tokens immediately following \texttt{[RingX]} will be interpreted as a number $N$, and we will count backwards $N$ atoms in placement order to determine the target of the ring bond. For example, \texttt{[Ring2]} indicates that the next \texttt{2} tokens will be interpreted as a \texttt{2}-digit base-16 number $N$. Ring bonds are stored until after all tokens have been read by the decoder; only then are ring bonds placed, and only if it would not violate valency. Due to the addition of groups, it is sometimes necessary to form ring bonds to atoms that are added after the current atom (e.g. ring bonds within groups). To indicate this, we use the \texttt{[->]} token before the number token to specify that we will count forwards instead of backwards.
\end{itemize}

All tokens can be modified by adding \texttt{=}, \texttt{\#}, \texttt{\textbackslash} or \texttt{/} to change the bond order or stereochemistry of their parent bond (e.g. \texttt{[\#Branch]} or \texttt{[/C]}). The parent bond is the bond to the previous atom. 

\subsection{Groups}

The primary difference between SELFIES and Group SELFIES is the addition of groups. Each group is defined as a set of atoms and bonds representing the molecular group with its \textit{attachment points}, indicating how the group can participate in bonding. Each attachment point has a specified maximum valency, which allows us to continue tracking available valency while decoding. These attachment points are labeled by \textit{attachment indices}, and the encoder and decoder will navigate around these attachment indices as described in \Cref{encode-decode}.

Users must specify the groups they want to use using a dictionary that maps group names to groups. This tells the encoder what groups to recognize, and tells the decoder how to map group tokens to groups. We call this dictionary a ``group set'', and every group set defines its own distinct instance of Group SELFIES. In particular, the decoder will not recognize a Group SELFIES string that contains group tokens not present in the current group set.

To distinguish group tokens from other tokens, we include a \texttt{:} character at the front of the token (e.g. \texttt{[:4benzene]}). All group tokens are of the form \texttt{[:S<group-name>]}, where \texttt{S} is the starting attachment index of the group, and \texttt{<group-name>} is any alphanumeric string that does not start with a number.

Optionally, a priority value can be specified for each group, indicating the priority with which the group should be recognized when encoding into Group SELFIES. Priority affects the Group SELFIES encoder as described in \Cref{encode-decode}. For each group, one can also specify its index overload value, which is the value the group token takes when the decoder must interpret the token as a number.

\begin{figure}[H]
  \centering
  \includegraphics[width=0.3\textwidth]{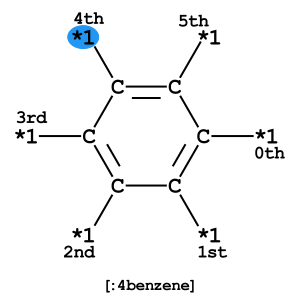}
  \caption{Representation of a possible ``benzene'' group and its use in a corresponding group token. The \texttt{*N} notation represents an attachment point with valency \texttt{N}. Each attachment point is labeled with its attachment index \texttt{0th,1st...} The starting attachment point represented in the token is also highlighted.}
  \label{fig:benzene-group}
\end{figure}

\subsection{Encoding and Decoding}
\label{encode-decode}
\textbf{Encoding:} To encode a molecule in Group SELFIES, the encoder first recognizes and replaces substructure matches of groups from the molecule. By default, the encoder iterates through the group set and recognizes the largest groups first, but users can override this by specifying a priority for each group. Setting a high priority value for a group indicates that it will be recognized first when encoding into Group SELFIES, ensuring that other group replacements will not overlap with this group. This encoding strategy iterates implies that increasing the size of the group set will increase the running time of the encoder linearly. We then traverse the graph similar to the encoding process for SMILES and SELFIES, while also placing the correct tokens for tracking the attachment indices of where the encoder entered and exited a group.

\textbf{Decoding:} When decoding Group SELFIES, the process is essentially the same as regular SELFIES except when reading group tokens. When a group token is read by the decoder, the group set dictionary determines the corresponding group. Subsequently, all atoms of the group are placed and the main chain is connected to the starting attachment point. The decoder selects the next attachment point to branch off from by reading in the next token as a \textit{relative index}. By adding the current attachment index to a relative index modulo the total number of attachment points in the group, the decoder selects the next attachment point. From the specified attachment point, the decoder implicitly branches off of the group and continues traversing until a \texttt{[pop]} token is read.
Once the branch is ``popped'', the decoder returns to the group and can navigate to the next attachment point using another relative index. If the selected attachment point is occupied, then the next available attachment point is used. If all attachment points have been used up, then the group itself is immediately ``popped'', returning to the most recent branchpoint before the group was placed.

We verified the robustness of Group SELFIES by encoding and decoding 25M molecules from the eMolecules database \cite{emolecules}. We provide a detailed example of encoding/decoding the molecule celecoxib in Appendix \ref{example-encode-decode}.

Group SELFIES manages chirality differently than SMILES and SELFIES. Rather than use @-notation to specify tetrahedral chirality, all chiral centers must be specified as groups. We provide an ``essential set'' of 23 groups which encode all relevant chiral centers in the eMolecules database. Equipped with this essential set, every molecule can be encoded-decoded while maintaining chirality. It is also an option to not use the essential set, or only use a subset of it, depending on what chiral centers are relevant to the problem at hand. If a molecule has a chiral center not specified in the group set, then encode-decode will not preserve chirality.

\subsection{Determining Fragments}
\label{fragments}

Group SELFIES has a built-in flexibility for assigning the set of fragments that make up a group set. Hence, the construction of a useful group set often remains an open design choice. Users can specify groups using a SMILES-like syntax, which could be useful if one knows what groups are synthetically available or are expected to be useful for their particular design task. Fragments can also be obtained from several fragment libraries found in the literature \cite{zdrazil2017rise, ertl2019systematic, Sharif_Global-Chem_A_Chemical_2022}. Generally, a useful set of groups will appear in many molecules in the dataset and replace many atoms, with similar fragments merged together to reduce redundancy.

In our experiments, we also tested various fragmentation algorithms that extract fragments from a dataset, including a naïve technique that cleaves side chains from rings and a method based on matched molecular pair analysis \cite{hussain2010computationally}. Several other fragmentation algorithms from cheminformatics can be readily applied \cite{sheng2013fragment, liu2017break, kutchukian2015fragment, muller2019flexible, degen2008art, guo2022data} and the essential features of Group SELFIES outlined in \Cref{sec:intro} remain robust to different fragment discovery methods.

\section{Experiments} \label{sec:experiments}
The experiments in the subsequent sections outline some of the advantages of Group SELFIES compared to SMILES and regular SELFIES representations. Concretely, we show that: (1) Group SELFIES is shorter and more compressible than SMILES and SELFIES; (2) Group SELFIES preserves useful properties during generation; (3) Group SELFIES improves distribution learning.

\subsection{Compactness}
\label{shorter}
\begin{figure}
  \centering
  \includegraphics[width=0.9\textwidth]{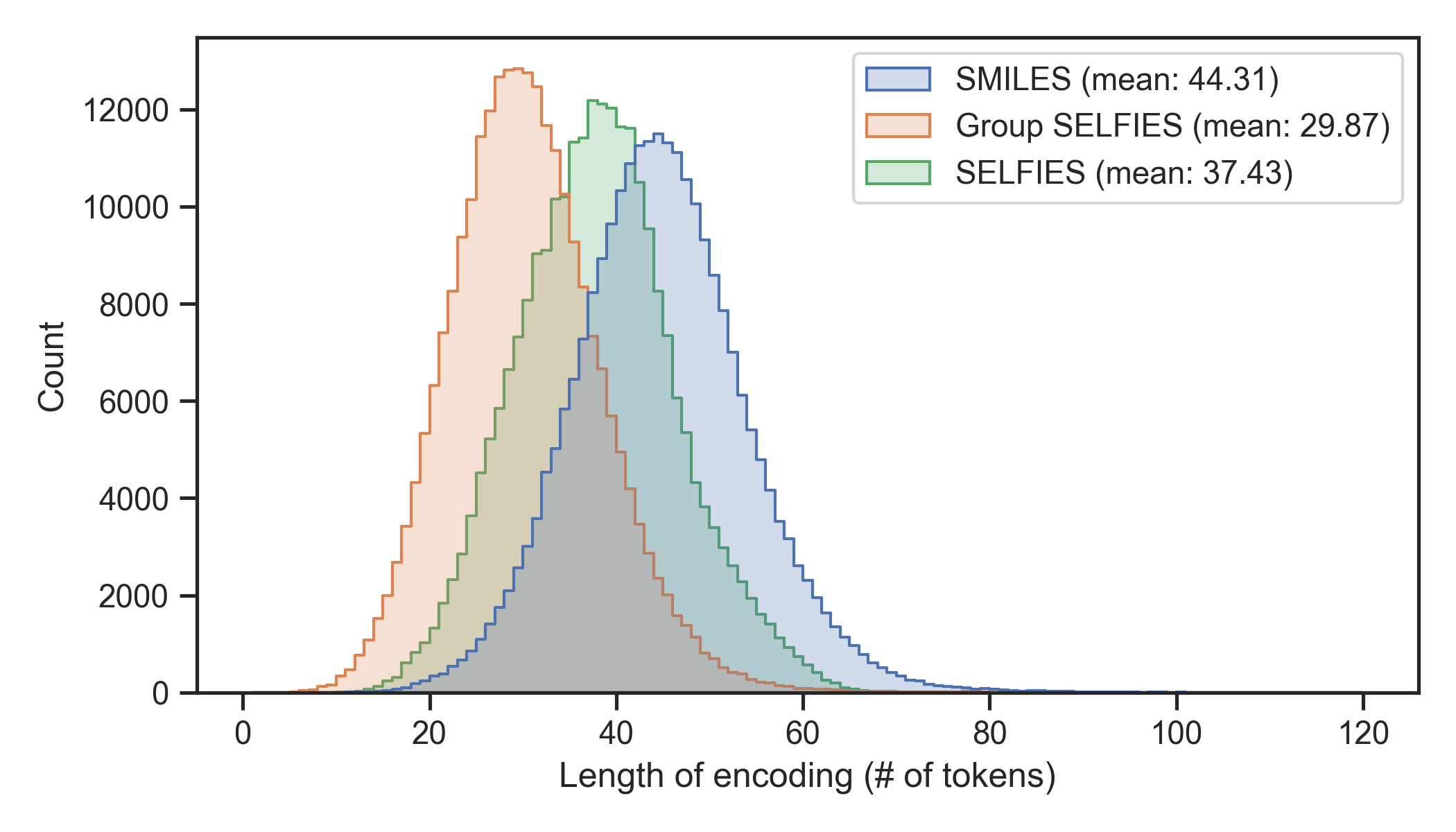}
  \caption{Histogram of lengths of SMILES, SELFIES, and Group SELFIES strings of the ZINC-250k dataset. Here, Group SELFIES uses a group set of 53 groups.}
  \label{fig:length_histogram}
\end{figure}

Group SELFIES strings are typically shorter than their SMILES and SELFIES equivalents when using a generic set of groups. In Figure \ref{fig:length_histogram}, this generic set was generated by taking a random selection of 10,000 molecules from ZINC-250k \cite{irwin2020zinc20} and fragmenting them into 30 useful groups using various algorithms (see \nameref{fragments}). We then combined these 30 groups with the 23 groups of the essential set. Figure \ref{fig:length_histogram} shows histograms of the lengths of SMILES/SELFIES/Group SELFIES strings of the entire ZINC-250k dataset. Length is the number of characters in SMILES strings, and the number of tokens in (Group) SELFIES strings. Group SELFIES strings are usually shorter than their SELFIES and SMILES counterparts because group tokens can represent multiple atoms in a molecule.

Since Group SELFIES has a larger alphabet than SMILES or SELFIES, we estimate the complexity of each representation with the compressed filesize of ZINC-250k. We find that out of all representations, Group SELFIES can be compressed the most (see Appendix \ref{compression}).

\subsection{Molecular Generation}
\label{primitive_gen}

\renewcommand{\algorithmiccomment}[1]{\# #1}
\begin{algorithm}
\caption{Sampling random (Group) SELFIES strings}
\begin{algorithmic}[1]
\STATE \textbf{given} molecule dataset $D$, and \textbf{encoder}, \textbf{decoder} from SELFIES or Group SELFIES

\STATE $S \leftarrow \{$\textbf{encoder}$($molecule$)$ \textbf{for} molecule $\in D\}$ \hfill \COMMENT{(Group) SELFIES strings}
\STATE $B \leftarrow$ concatenate$\{$tokenize$(s)$ \textbf{for} $s \in S\}$ \hfill \COMMENT{bag of tokens}

\REPEAT
\STATE \textbf{select} random string $s \in S$
\STATE length $l \leftarrow$ length$(s)$
\STATE tokens $\leftarrow$ \textbf{sample} $l$ tokens from $B$
\STATE string $\leftarrow$ concatenate$($tokens$)$
\STATE \textbf{output} \textbf{decoder}$($string$)$ \hfill \COMMENT{generated molecule}
\UNTIL{enough molecules are generated}

\end{algorithmic}
\end{algorithm}

To specifically compare the suitability of the SELFIES and Group SELFIES representations for molecular generation, we use a primitive generative model which samples random strings. First, we convert a subset of $N=$ 100,000 molecules from ZINC-250k into (Group) SELFIES strings. Then, we tokenize all strings and combine them into a single bag of tokens. To generate a new string, we first pick a random (Group) SELFIES string from our chosen subset and take its length $l$. We then randomly sample $l$ tokens from the bag, and concatenate into a generated string. We generate $N$ random strings for each representation. For Group SELFIES, we use the same 53 groups used for the length histogram in \Cref{shorter}.

\begin{figure}[H]
    \centering
    \includegraphics[width=\textwidth]{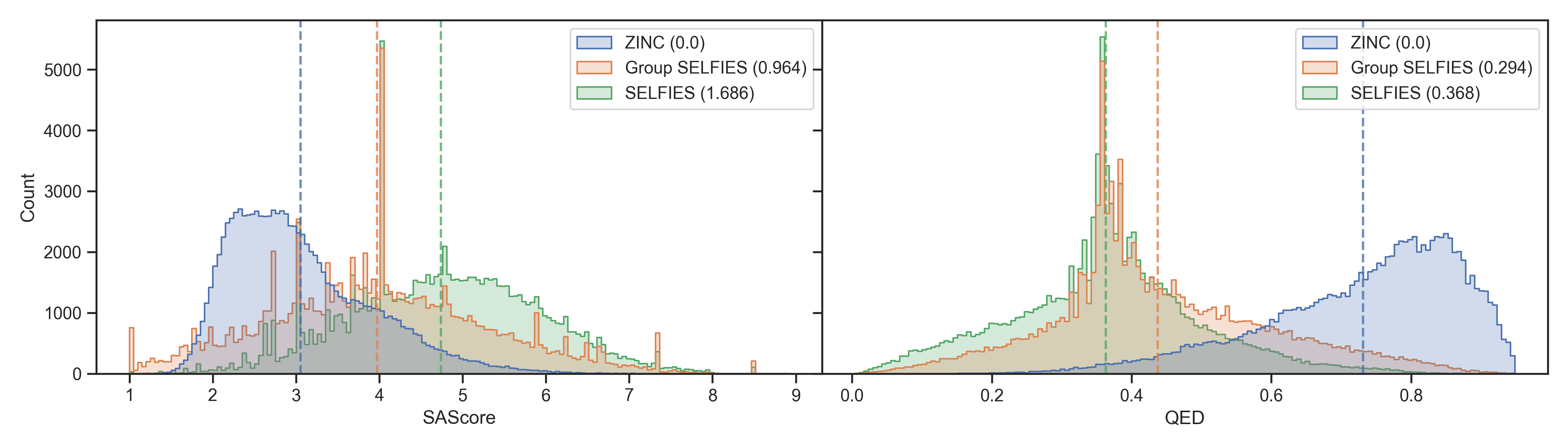}
    \caption{Molecules generated by our primitive generative model are binned by SAScore and QED. For both properties, generated Group SELFIES have greater overlap with the original ZINC distribution. Bracketed values indicate the Wasserstein distance (a measure of overlap) to the ZINC distribution. Dashed lines indicate the means.}
    \label{fig:generated_zinc}
\end{figure}

We show histograms of the SAScore \cite{ertl2009estimation} and QED \cite{bickerton2012quantifying} of molecules generated from ZINC in Figure \ref{fig:generated_zinc}. The distributions of generated Group SELFIES more closely overlap with the original ZINC dataset than the generated SELFIES, showing that even with an extremely simplistic generative model, Group SELFIES can preserve important structural information. We perform a similar analysis for a dataset of nonfullerene acceptors (NFA) \cite{lopez2017design} in Appendix \ref{nfa} and find that Group SELFIES preserves many aromatic rings in contrast to SELFIES, which rarely ever preserves aromatic rings.

\subsection{Distribution Learning}

To further quantify the effectiveness of Group SELFIES in generative models, we use the MOSES benchmarking framework \cite{polykovskiy2018moses} to evaluate variational autoencoders (VAEs) trained with both Group SELFIES and SELFIES strings. Models were trained for 125 epochs. The group set for the Group SELFIES VAE was created by fragmenting the training set provided by MOSES and selecting the 300 most diverse groups. A set of 100,000 molecules was then generated from each model and evaluated on the metrics provided by MOSES.

\begin{table}[ht]
\centering
\scalebox{.7}{
\begin{tabular}{llllllll}
\toprule
\multirow{2}{*}{Model}  & \multirow{2}{*}{Valid ($\uparrow$)}  & \multirow{2}{*}{Unique@1k ($\uparrow$)}  & \multirow{2}{*}{Unique@10k ($\uparrow$)}  & \multicolumn{2}{c}{FCD ($\downarrow$)} & \multicolumn{2}{c}{SNN ($\uparrow$)}\\
 &  &  &  & Test & TestSF & Test & TestSF\\
\midrule
 {\it Train } &      {\it 1.0 } &      {\it 1.0 } &            {\it 1.0 } &          {\it 0.008 } &        {\it 0.4755 } &         {\it 0.6419 } &         {\it 0.5859 } \\
     Group-VAE-125 &  {\bf 1.0(0) } &  {\bf 1.0(0) } &              0.9985(4) &  {\bf 0.1787(29) } &  {\bf 0.734(109) } &  {\bf 0.6051(4) } &  {\bf 0.5599(3) } \\
   SELFIES-VAE-125 &  {\bf 1.0(0) } &        0.9996(5) &  {\bf 0.9986(4) } &              0.6351(43) &             1.3136(128) &              0.6014(3) &              0.5566(2) \\
\end{tabular}
}
\scalebox{.7}{
\begin{tabular}{lllllllll}
\toprule
\multirow{2}{*}{Model}  & \multicolumn{2}{c}{Frag ($\uparrow$)} & \multicolumn{2}{c}{Scaf ($\uparrow$)} & \multirow{2}{*}{IntDiv ($\uparrow$)}  & \multirow{2}{*}{IntDiv2 ($\uparrow$)}  & \multirow{2}{*}{Filters ($\uparrow$)}  & \multirow{2}{*}{Novelty ($\uparrow$)} \\
 & Test & TestSF & Test & TestSF &  &  &  & \\
\midrule
 {\it Train } &         {\it 1.0 } &         {\it 0.9986 } &         {\it 0.9907 } &            {\it 0.0 } &         {\it 0.8567 } &         {\it 0.8508 } &            {\it 1.0 } &            {\it 1.0 } \\
     Group-VAE-125 &  {\bf 0.9995(0) } &  {\bf 0.9977(1) } &  {\bf 0.9649(21) } &              0.0608(65) &  {\bf 0.8587(1) } &  {\bf 0.8528(1) } &  {\bf 0.9623(7) } &              0.7187(11) \\
   SELFIES-VAE-125 &              0.9989(0) &              0.9965(1) &              0.9588(15) &  {\bf 0.0675(37) } &              0.8579(1) &              0.8519(1) &                0.96(4) &  {\bf 0.7345(16) } \\
\bottomrule
\end{tabular}
}

\center
\caption{Group SELFIES VAE and SELFIES VAE evaluated on MOSES metrics. The Group SELFIES VAE mostly matches or outperforms the SELFIES VAE.}
\end{table}

For most metrics, Group SELFIES performs approximately the same as SELFIES. Validity is the percentage of generated molecules that are accepted by RDKit's parser. Uniqueness is the percentage of generated molecules that are not identical to any other generated molecule. Similarity to nearest neighbor (SNN) is the average Tanimoto similarity between generated molecules and the nearest neighbor in the reference set. Fragment similarity (Frag) is a cosine similarity based on the distribution of BRICS fragments \cite{degen2008art} of generated and reference molecules. Scaffold similarity (Scaf) is a cosine similarity based on the distribution of Bemis-Murcko scaffolds \cite{murcko1996properties} of generated and reference molecules. Internal diversity (IntDiv) measures the chemical diversity of the generated molecules using Tanimoto similarity. Filters is the fraction of generated molecules that pass filters for unwanted fragments. Novelty is the fraction of generated molecules not in the training set.

\begin{figure}[H]
     \centering
     \includegraphics[width=0.49\textwidth]{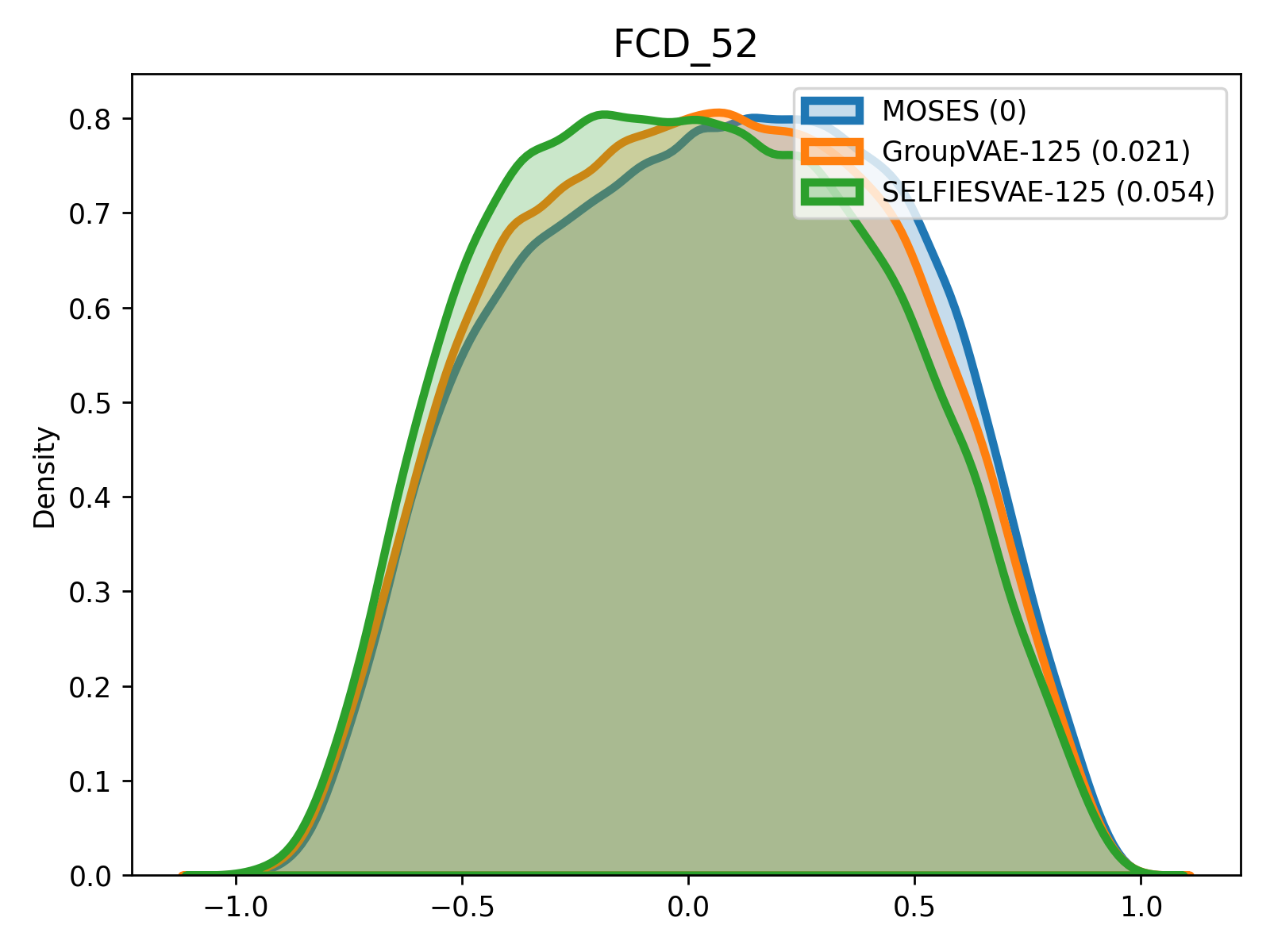}
     \includegraphics[width=0.49\textwidth]{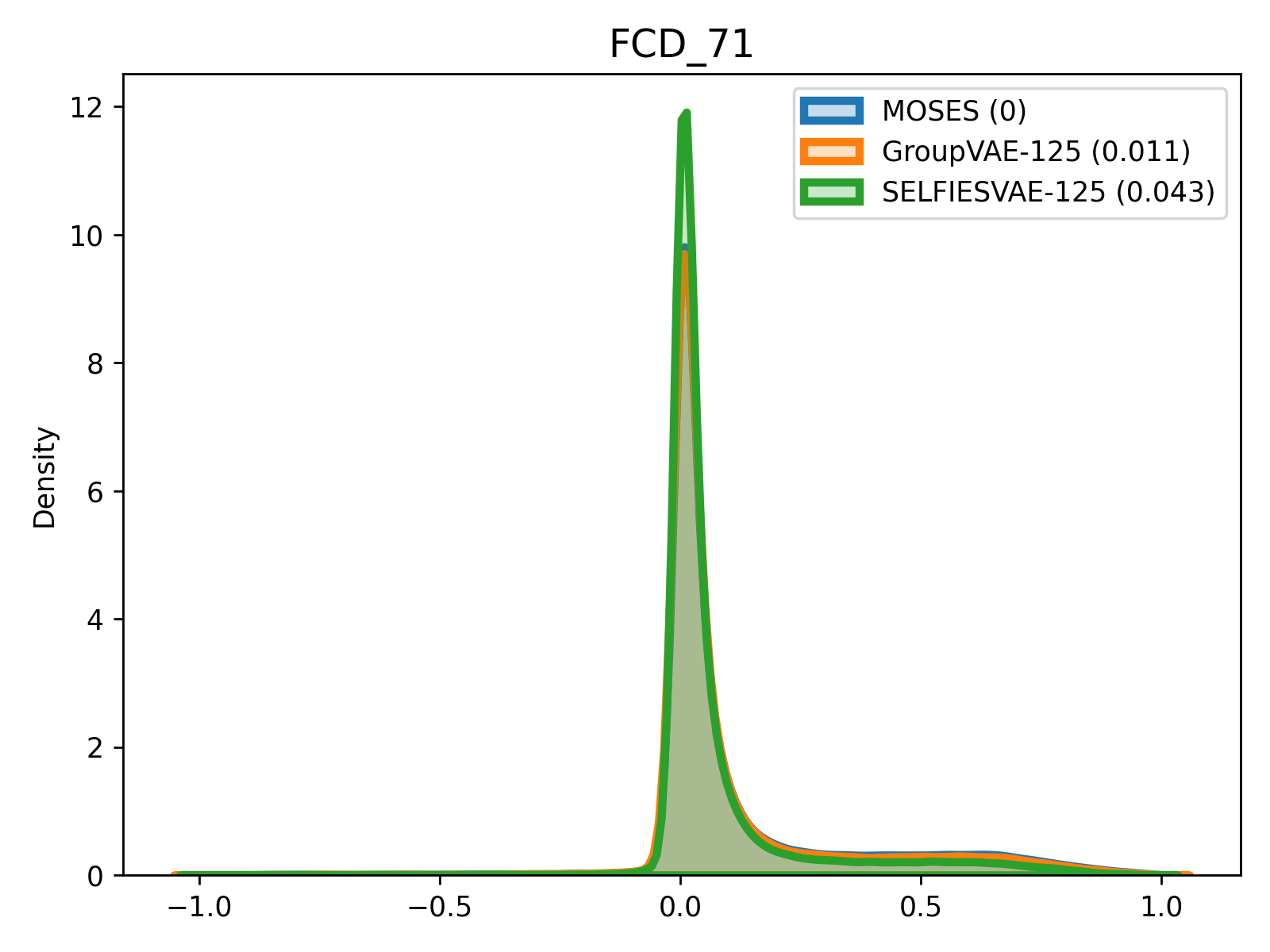}
     \\
     \includegraphics[width=0.49\textwidth]{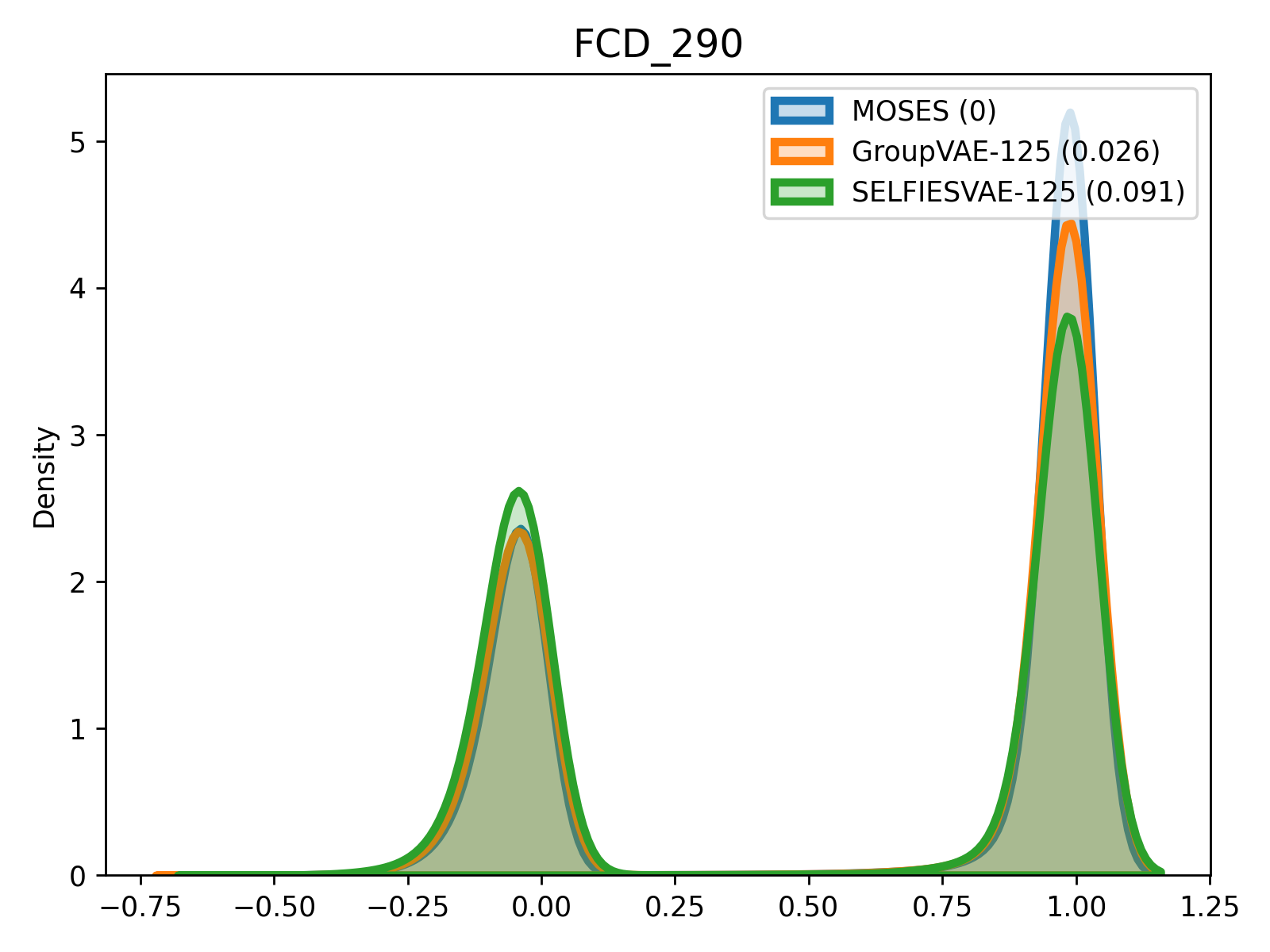}
     \includegraphics[width=0.49\textwidth]{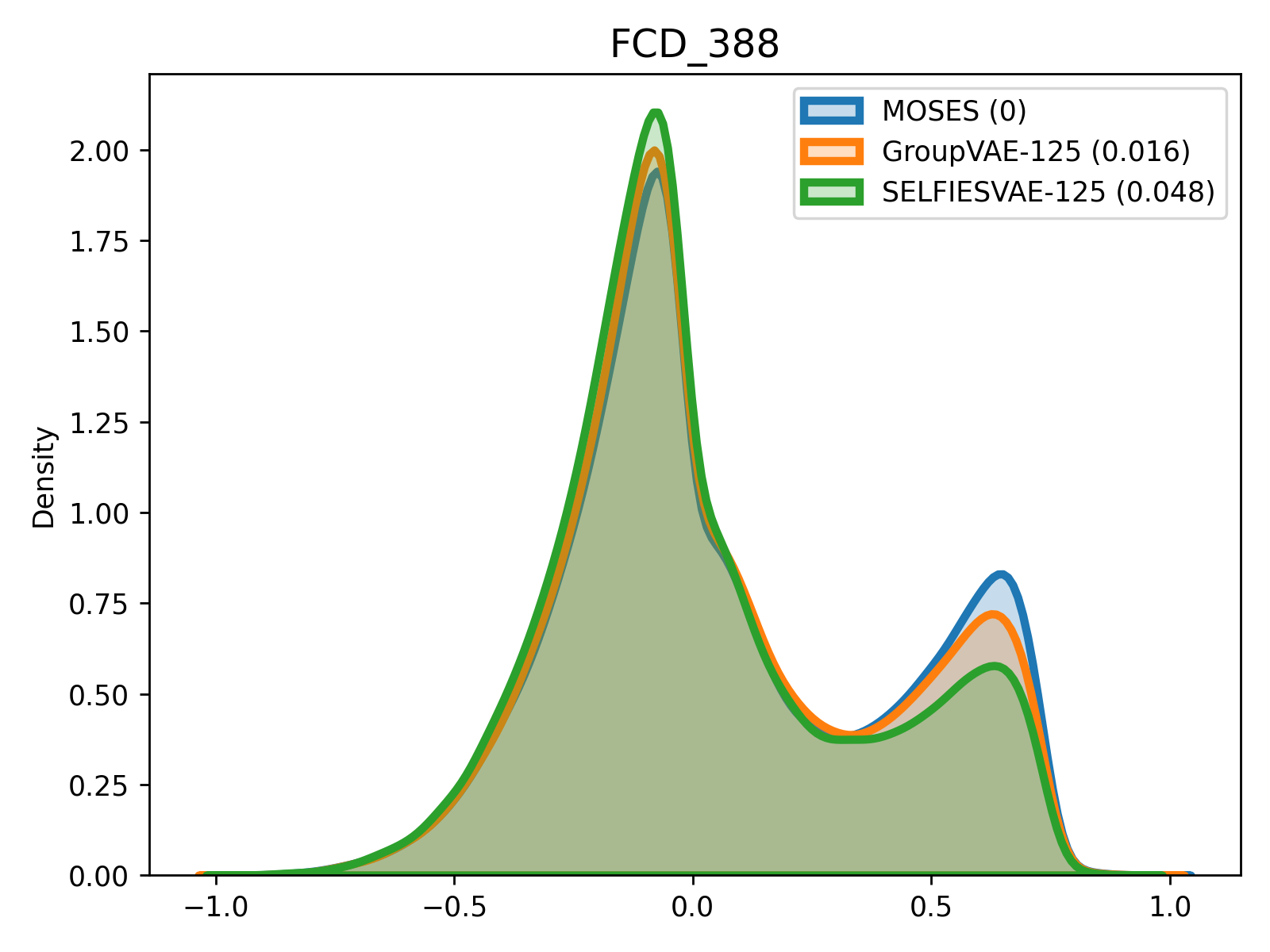}
    \caption{Distribution of some values from the second-to-last layer of ChemNet for molecules generated by Group SELFIES and SELFIES compared to the validation set. The difference in distributions is used to calculate FCD. Bracketed values in the legend represent the Wasserstein distance to the original MOSES distribution.}
    \label{fig:vae-distributions}
\end{figure}

The Group SELFIES model performs especially well on the Fréchet ChemNet Distance (FCD) metric \cite{preuer_frechet_2018} when compared to SELFIES. FCD measures the difference between the activations of the penultimate layer of ChemNet (a model trained to predict the bioactivity of molecules) in the validation set and in the generated set. Due to how ChemNet was trained, the activations are likely to encode a mixture of biological and chemical properties important to drug likelihood. This makes comparing these activations more informative than comparing standard properties like logP or molecular weight, where the correlation to bioactivity is weaker and less deliberate. To visualize FCD, some indices of the penultimate activations of ChemNet are graphed in Figure \ref{fig:vae-distributions}. Generated Group SELFIES match these distributions more closely than generated SELFIES.

\section{Discussion} \label{sec:discussion}

Our experiments show that Group SELFIES has noticeable advantages compared to SMILES and SELFIES representations, including greater readability provided by the group tokens. With regards to SMILES, the primary advantage is chemical robustness. The comparison with SELFIES is more nuanced, as discussed in the section below.

\subsection{Group SELFIES vs SELFIES}

\textbf{Substructure Control:} Group SELFIES provides more fine-grained control of substructures, which creates the following advantages: (1) An important scaffold can be preserved during optimization; (2) Chiral and charged groups can be preserved during optimization, ensuring that charged tokens do not proliferate and create radicals; (3) Synthetically accessible building blocks can be chosen as groups to improve synthesizability.

\textbf{Substructure Control with SELFIES:} Various techniques applied to SELFIES can mitigate the challenges of preserving structure. One such example is to simply combine substrings of SELFIES strings together. Indeed, further experiments in Appendix \ref{substrings} show that simply replacing all group tokens by their SELFIES substrings shows similar performance to Group SELFIES. Within the SELFIES framework, however, an inserted substring will not necessarily have that exact substructure when decoded because the first token of the inserted substring may need to be interpreted as a number, which can have cascading effects for the rest of the substring. It is also likely that upon further insertions, that substructure will not be preserved. Additionally, it is also not clear how an insertion based approach can create groups with 3 or more branches, since creating a third branch requires insertion in the middle of the group substring. 

\begin{figure}[H]
     \centering
     \includegraphics[width=\textwidth]{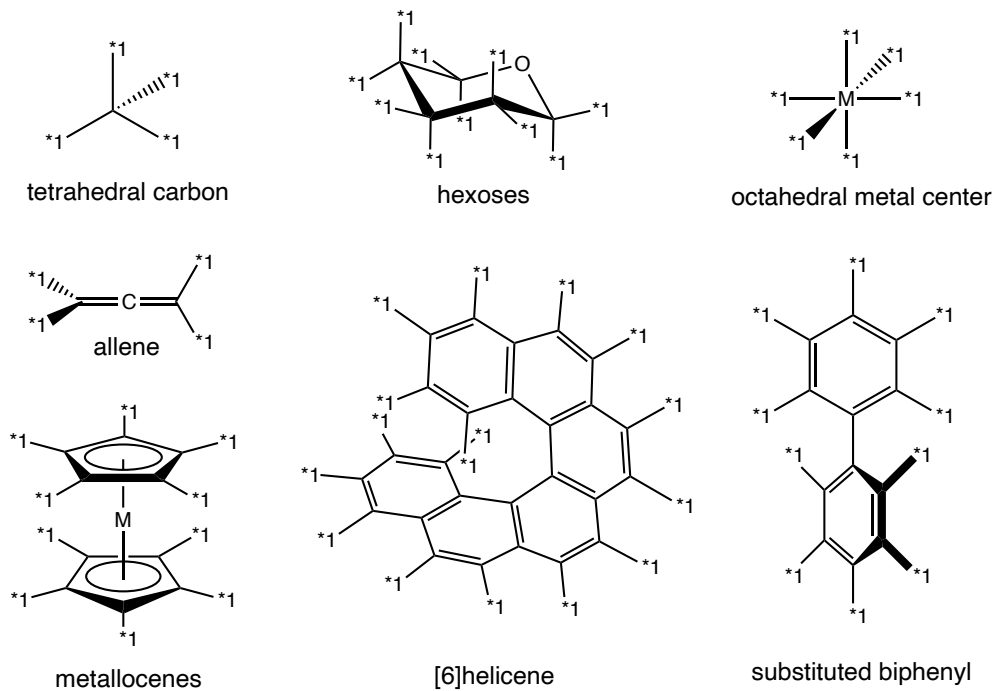}
    \caption{Examples of chiral groups that can be represented in group SELFIES. Tetrahedral carbon, hexoses, and octahedral metal centers have local chirality, whereas allenes, metallocenes, helicenes, and substituted biphenyls have global chirality.}
    \label{fig:chiral-groups}
\end{figure}

\textbf{Extended Chirality:} 
Group SELFIES is theoretically capable of representing molecules with extended chirality which are traditionally not able to be represented with SMILES and SELFIES. These representations can only handle \textit{local} chirality - that is, chirality with a single atom as the chiral center. This is in contrast to \textit{global} chirality, where there may be an axis or plane of chirality. Group SELFIES can handle global chirality by taking an entire complex or chiral substructure and abstracting it into a group, leaving attachment points on the outside for varying functionalization. Figure \ref{fig:chiral-groups} shows examples of groups with local and global chirality that Group SELFIES can handle. We leave the proper implementation of representing global chirality to future work.

\textbf{Computational Speed:} One tradeoff of Group SELFIES is that encoding and decoding is usually slower than with SELFIES, likely due to overhead of RDKit operations. The encoder is particularly slow as it relies on performing a substructure match for every group in the group set. The decoder is faster than the encoder, though still slower than the SELFIES decoder. See Appendix \ref{timing} for timing. To improve computational performance in future work, one could exploit substructure control of Group SELFIES to reduce the number of encode/decode calls needed to obtain high performers. Additionally, the speed of encoding and decoding operations can be improved with distributed computing, since Group SELFIES is trivially parallelizable for a fixed group set.

\subsection{Future Work}
One promising extension of Group SELFIES is to incorporate more flexibility into the representation. In such a case, a group token can represent an entire scaffold, except without the atom identities. Other tokens can then identify the atom types on the scaffolds. This would allow optimization of the atom types while maintaining the topological structure of the scaffold. Another current limitation of Group SELFIES is that groups cannot overlap; more work is needed to develop a representation that acknowledges how groups might overlap, particularly for generating polycyclic compounds. A sequence-based representation of cellular complexes \cite{hajij2020cell} or hypergraphs might suggest a promising direction.

Finally, given this paper's focus on the molecular representation, we only applied rather simple generative modeling methods. We hope that future work can leverage Group SELFIES to perform molecular generation with more advanced generative methods, including chemical language models, deep generative models, and evolutionary methods.

\begin{ack}
We thank Luca Thiede, Akshat Nigam, Robert Pollice, Kjell Jorner, Gary Tom, Nathanael Kusanda, Edwin Yu, Naruki Yoshikawa, and Felix Strieth-Kalthoff for useful discussions.

Calculations testing the robustness of Group SELFIES were performed on the Niagara supercomputer at the SciNet HPC Consortium. SciNet is funded by: the Canada Foundation for Innovation; the Government of Ontario; Ontario Research Fund - Research Excellence; and the University of Toronto.

This research was partially supported by funds from Intel Corporation.

A.A.-G. thanks Anders G. Frøseth for his generous support. A.A.-G. also acknowledges the generous support of Natural Resources Canada and the Canada 150 Research Chairs program.

\end{ack}

\bibliographystyle{unsrtnat}
\bibliography{refs}
\clearpage
\appendix
\section{Appendix}

\subsection{Example}
\label{example-encode-decode}

As an example, we will go through encoding and decoding the molecule celecoxib.

\begin{figure}[H]
  \centering
  \includegraphics[width=0.4\textwidth]{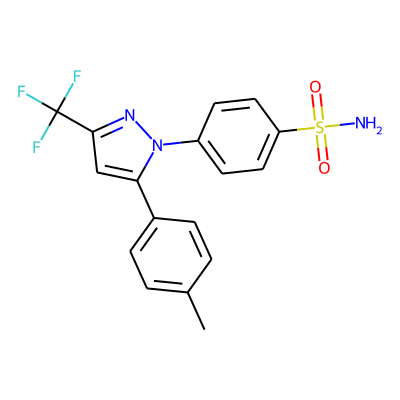}
  \caption{Celecoxib}
  \label{fig:celecoxib}
\end{figure}

We will define our group set as consisting of 4 main groups: trifluoromethane, toluene, benzene, and sulfonamide.

\begin{figure}[H]
  \centering
  \includegraphics[width=\textwidth]{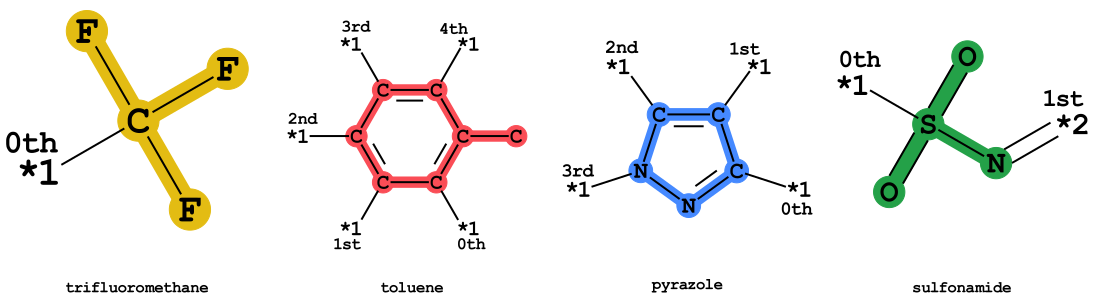}
  \caption{Group set used in this example.}
  \label{fig:example-grammar}
\end{figure}

We will now extract occurrences of the groups in the molecule. Since priorities were not specified, we match groups by size in descending order.

\begin{figure}[H]
  \centering
  \includegraphics[width=\textwidth]{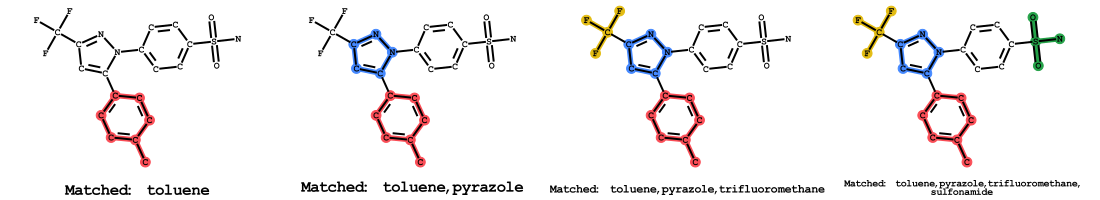}
  
\end{figure}

We will now traverse the graph. The following diagrams demonstrate this process.

\begin{figure}[H]
  \centering
  \includegraphics[width=\textwidth]{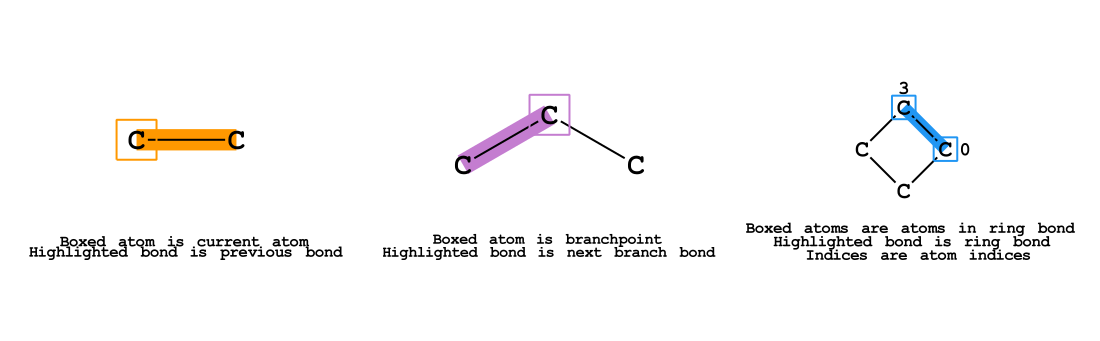}
  
\end{figure}

\begin{figure}[H]
  \centering
  \includegraphics[width=\textwidth]{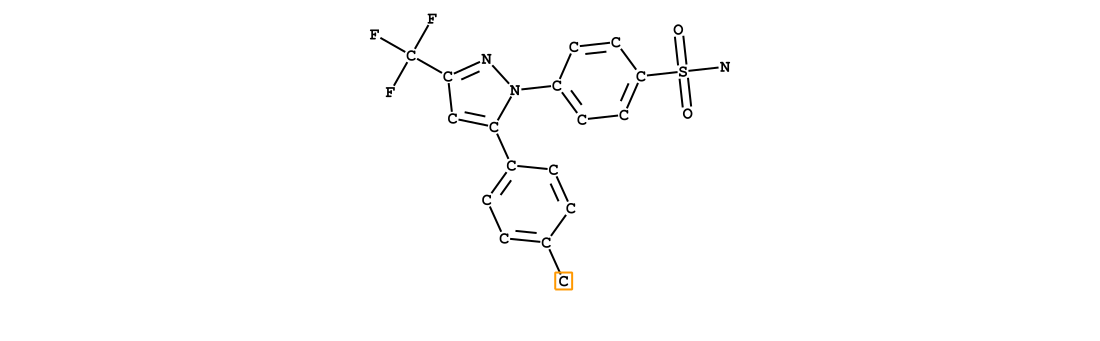}
  
\end{figure}

We then start traversing at an arbitrary atom, in this case the methyl carbon of the toluene.

\begin{figure}[H]
  \centering
  \includegraphics[width=\textwidth]{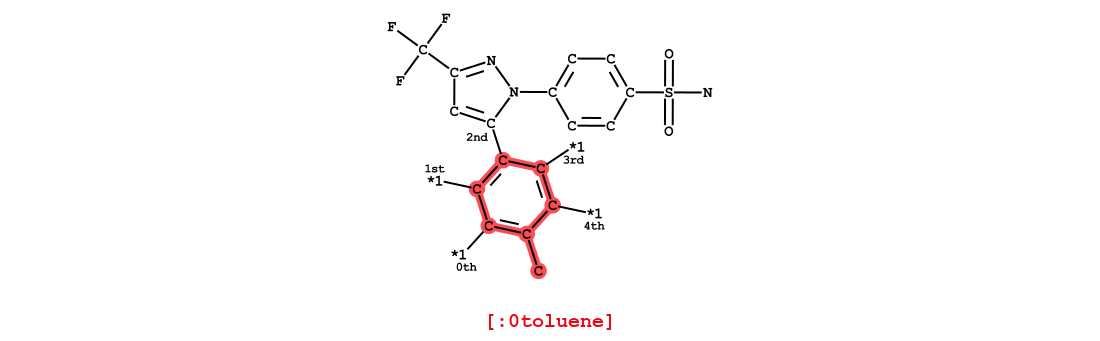}
  
\end{figure}

We find that this atom belongs to a group, and so place the appropriate group token. We arbitrarily select 0 to be the starting attachment point, since there is no previous connection into this group.

\begin{figure}[H]
  \centering
  \includegraphics[width=\textwidth]{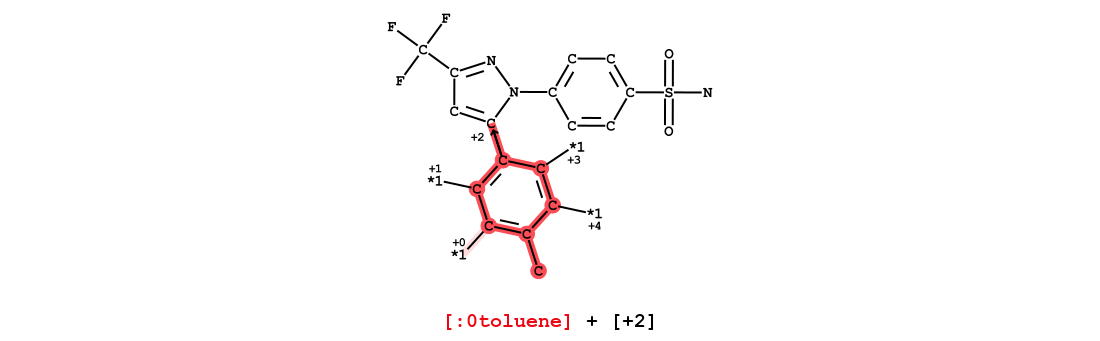}
  
\end{figure}

We then place an index token of \texttt{[+2]} to go from the \texttt{0th} attachment point to the \texttt{2nd} attachment point.  
For readability, index tokens are shown as numbers, but in the actual encoder output a token with the appropriate overload value would be used (e.g. in this case \texttt{[Ring2]} is overloaded to \texttt{[+2]}).
\begin{figure}[H]
  \centering
  \includegraphics[width=\textwidth]{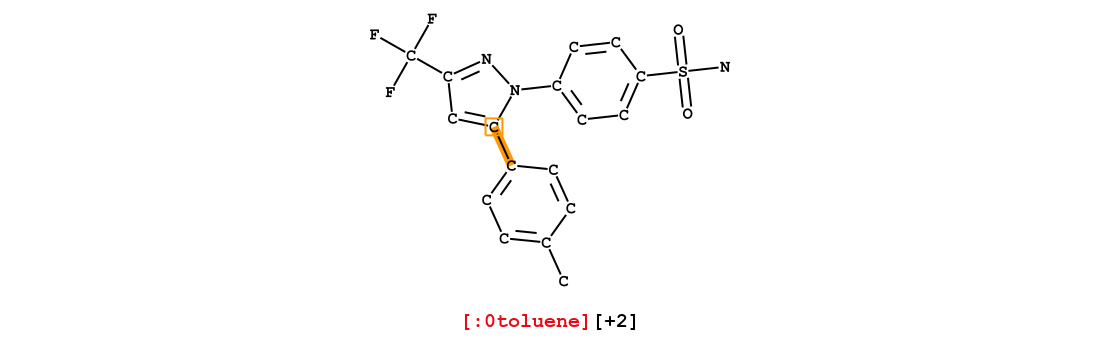}
  
\end{figure}
\begin{figure}[H]
  \centering
  \includegraphics[width=\textwidth]{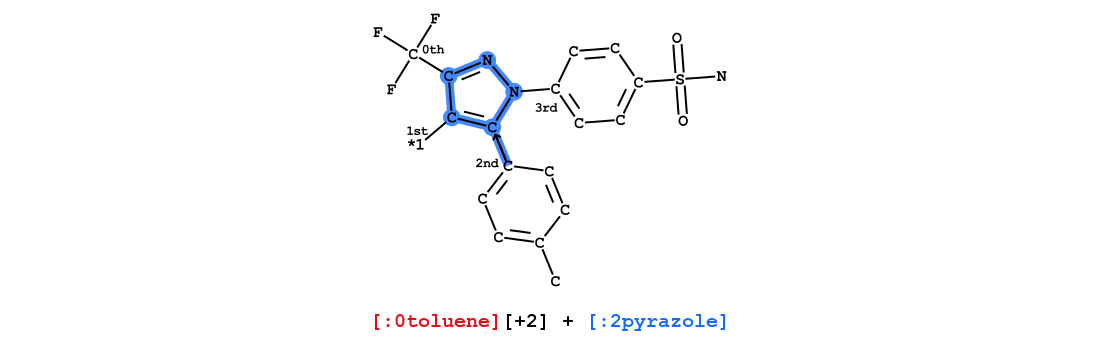}
  
\end{figure}

We again find that our current atom is in a group. This time, we place the group token with the starting attachment point at the \texttt{2nd} attachment index, since that is where the last bond entered from.

\begin{figure}[H]
  \centering
  \includegraphics[width=\textwidth]{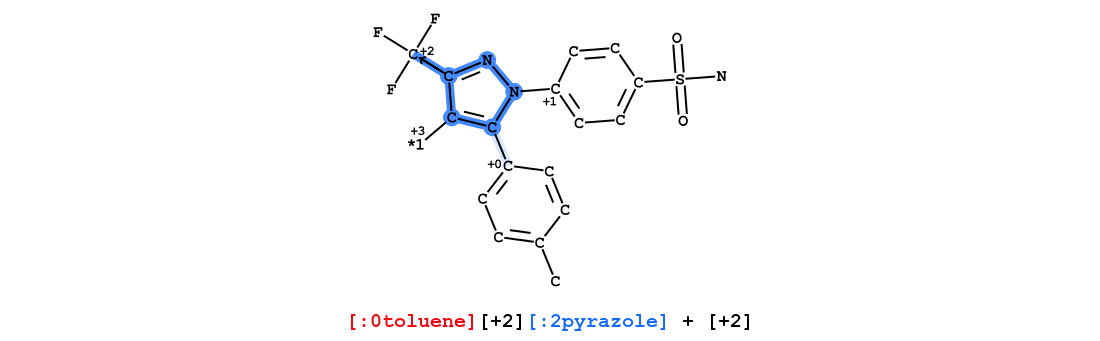}
  
\end{figure}

\begin{figure}[H]
  \centering
  \includegraphics[width=0.49\textwidth]{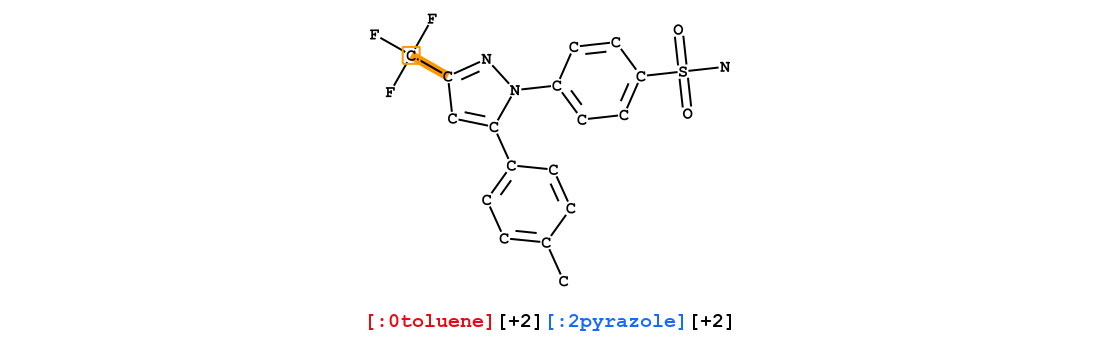}
  \includegraphics[width=0.49\textwidth]{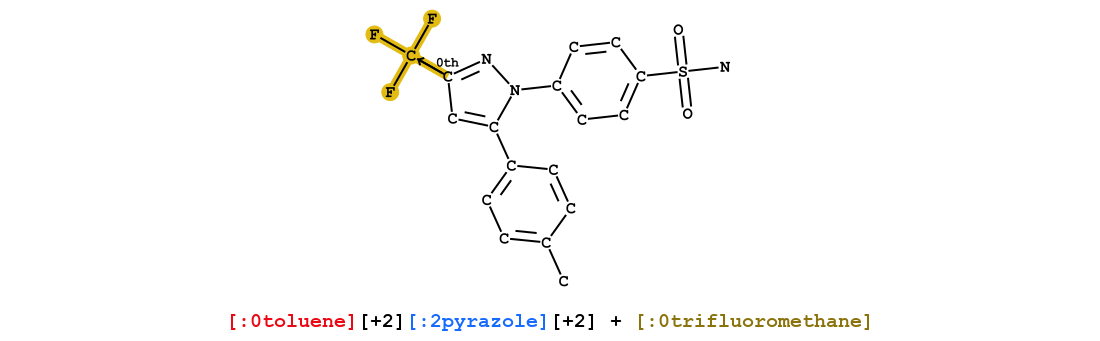}

\end{figure}

\begin{figure}[H]
    \centering
    \includegraphics[width=\textwidth]{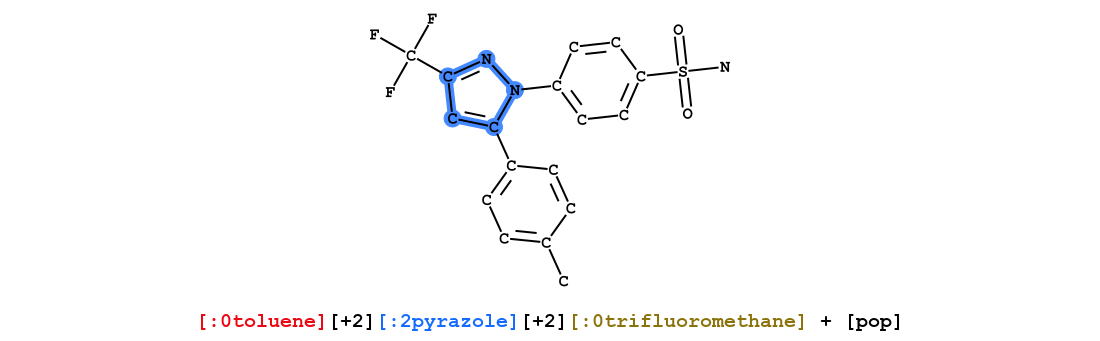}
\end{figure}

We then use the \texttt{[pop]} token to exit the current group. In this case, we go from the trifluoromethane group back to the pyrazole group. 

\begin{figure}[H]
  \centering
  \includegraphics[width=\textwidth]{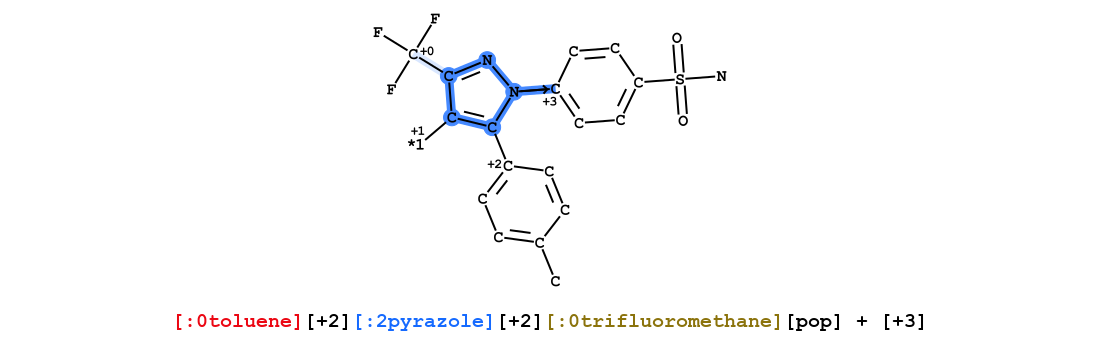}
  
\end{figure}

The next few steps build out the carbon chain of the benzene using atomic tokens, including placing a branch and a ring bond.
\begin{figure}[H]
  \centering
  \includegraphics[width=\textwidth]{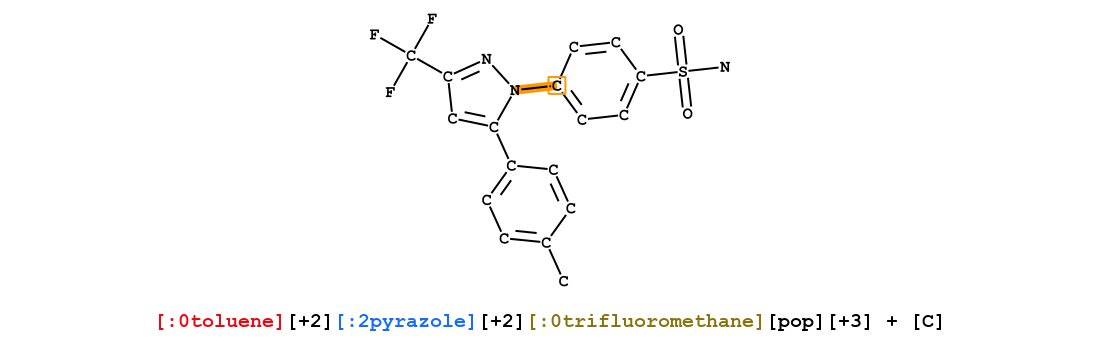}
  
\end{figure}
\begin{figure}[H]
  \centering
  \includegraphics[width=\textwidth]{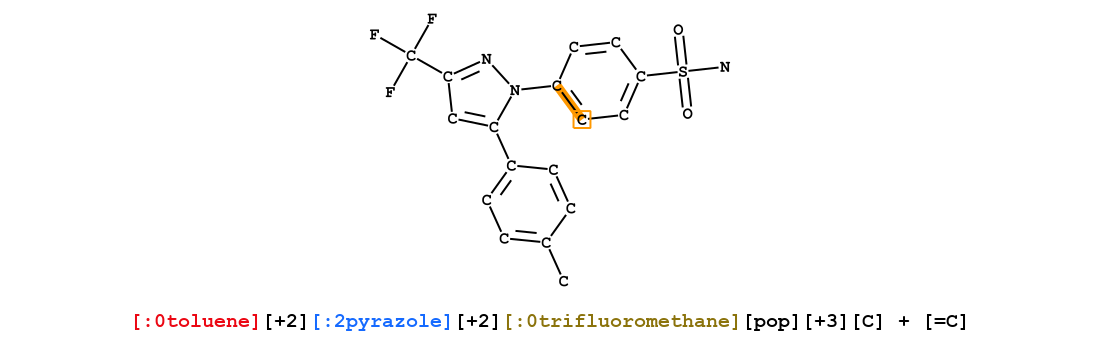}
  
\end{figure}
\begin{figure}[H]
  \centering
  \includegraphics[width=\textwidth]{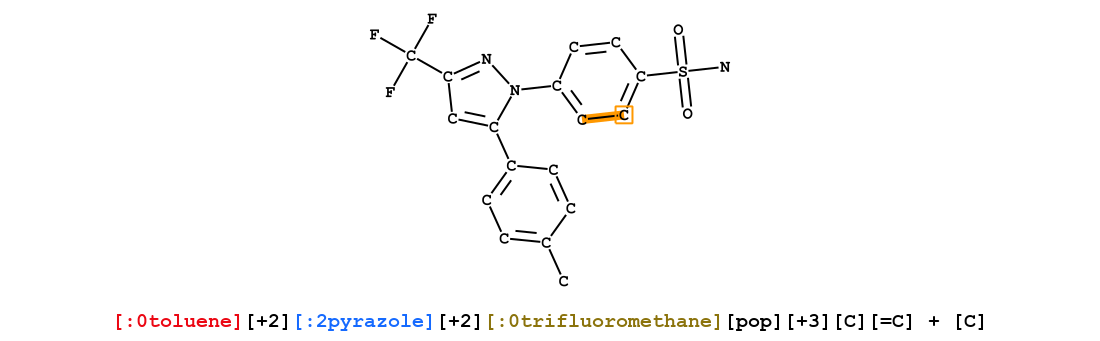}
  
\end{figure}
\begin{figure}[H]
  \centering
  \includegraphics[width=\textwidth]{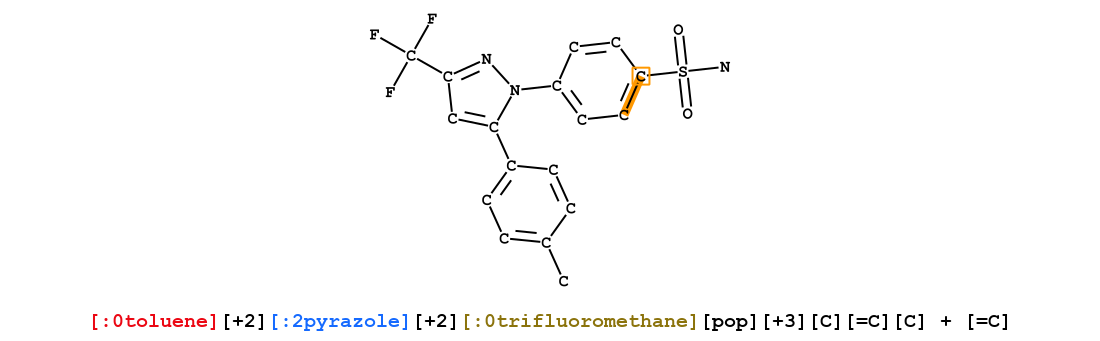}
  
\end{figure}
\begin{figure}[H]
  \centering
  \includegraphics[width=\textwidth]{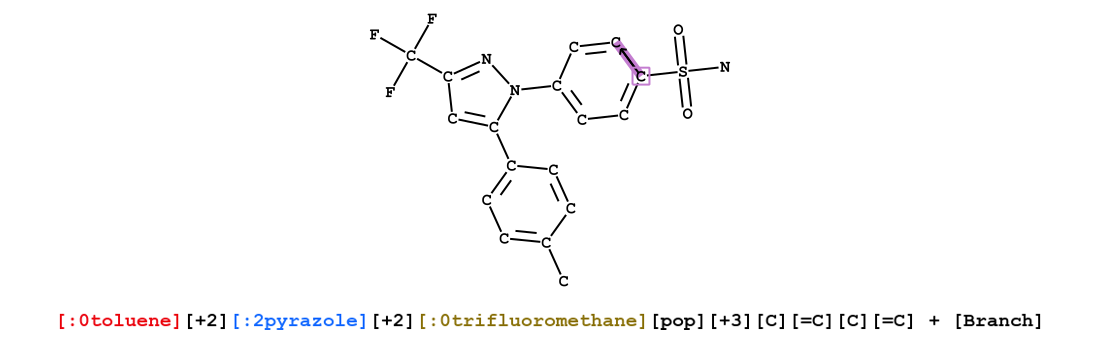}
  
\end{figure}
\begin{figure}[H]
  \centering
  \includegraphics[width=\textwidth]{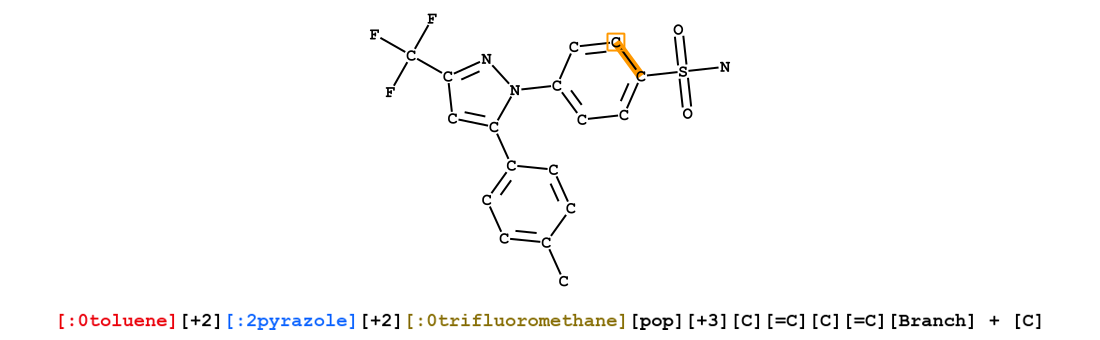}
  
\end{figure}
\begin{figure}[H]
  \centering
  \includegraphics[width=\textwidth]{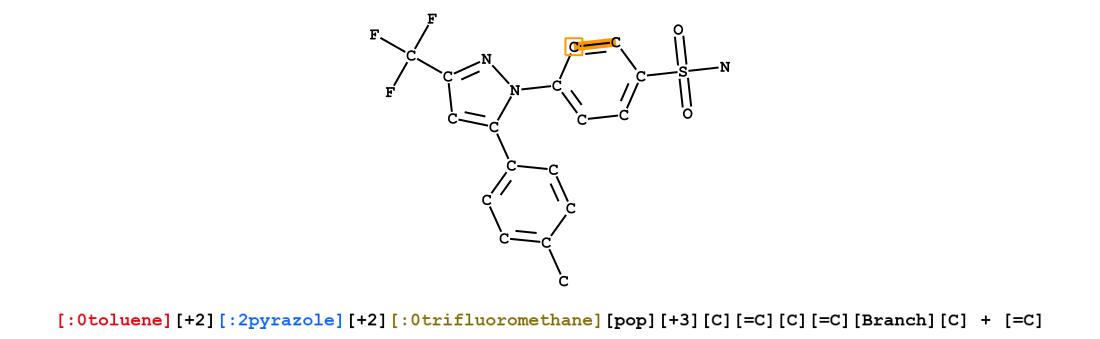}
  
\end{figure}
\begin{figure}[H]
  \centering
  \includegraphics[width=\textwidth]{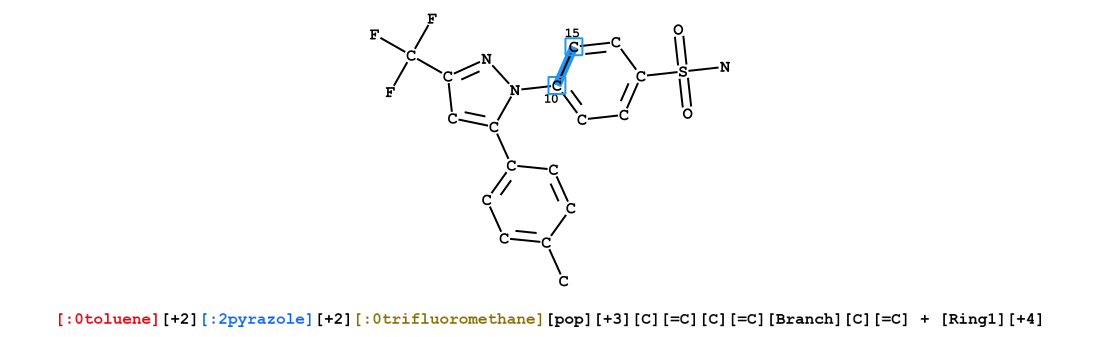}
  
\end{figure}
\begin{figure}[H]
  \centering
  \includegraphics[width=\textwidth]{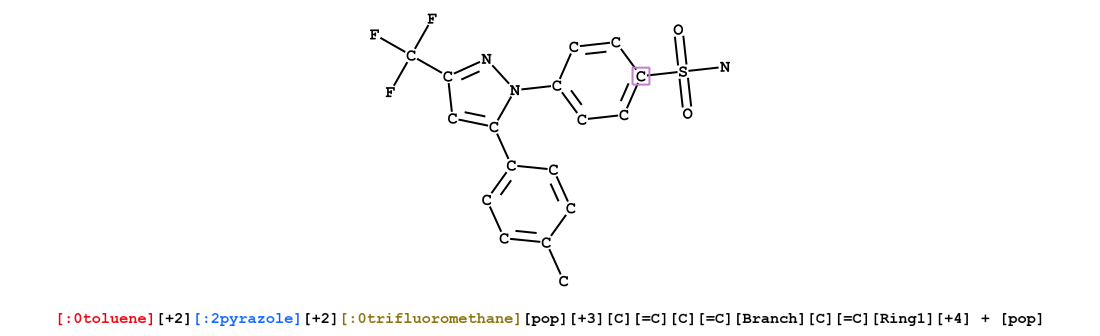}
  
\end{figure}

After the benzene is completed, the molecule is finished by placing a sulfonamide group.

\begin{figure}[H]
  \centering
  \includegraphics[width=\textwidth]{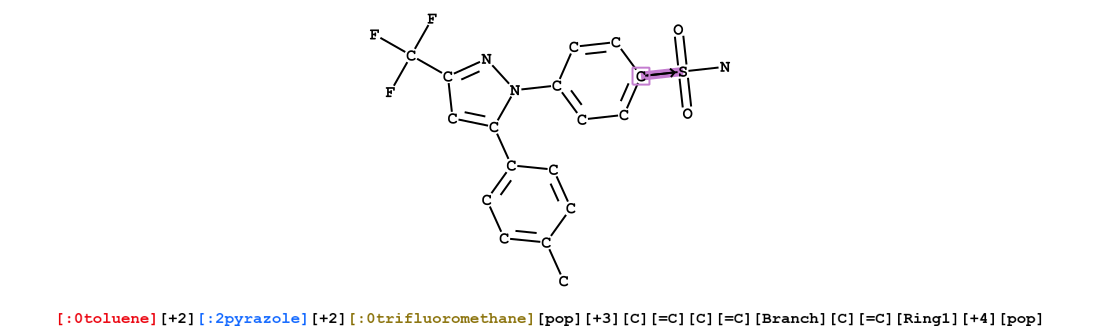}
  
\end{figure}
\begin{figure}[H]
  \centering
  \includegraphics[width=\textwidth]{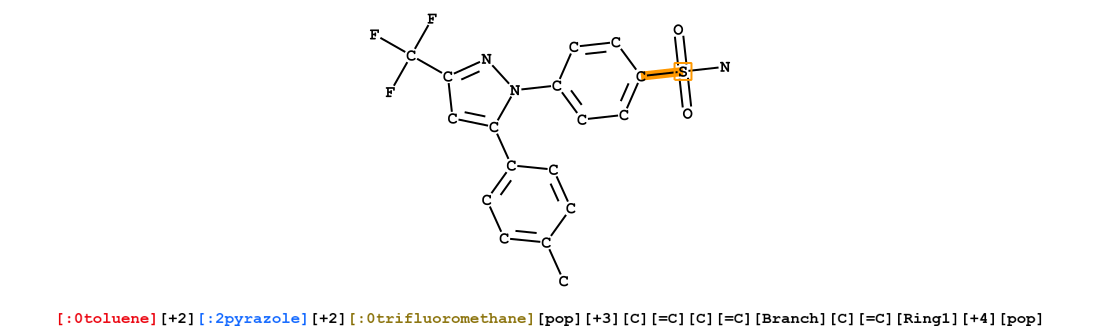}
  
\end{figure}
\begin{figure}[H]
  \centering
  \includegraphics[width=\textwidth]{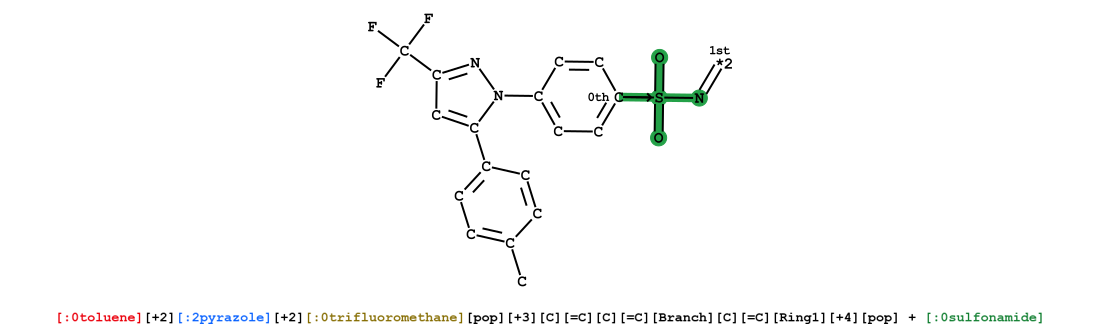}
  
\end{figure}

The decoding process for this Group SELFIES string will read in token-by-token, building the molecular graph in the same order as shown in this encoding example.

\subsection{Complexity and compression}

To take into account the alphabet size of SMILES, SELFIES, and Group SELFIES, we compress each representation using \texttt{zlib} to estimate the complexity of each representation. We first do index encoding to convert each string into a list of ints by replacing each unique token with a unique int. We then compress this representation using \texttt{zlib}.

\label{compression}
\begin{table}[h]
\begin{tabular}{ccccc}
\toprule
Representation  & $\#$ unique tokens  & strings  & +index encoding  & \begin{tabular}{c}+index encoding\\+\texttt{zlib} compression\end{tabular} \\
\midrule
SMILES & 34 & {\bf 11.80 MB}  & 23.11 MB  & 3.70 MB \\
SELFIES & 107 & 46.44 MB & 19.68 MB & 3.97 MB \\ 
Group SELFIES  & 247    & 42.05 MB & {\bf 15.90 MB}  & {\bf 3.61 MB} \\
\bottomrule
\end{tabular}
\label{table:sizes}
{\center 
\caption{Filesize of ZINC-250k when represented in SMILES, SELFIES, and Group SELFIES. When using strings, SMILES takes up the least space. When using index encoding and then compressing using \texttt{zlib}, Group SELFIES takes up the least space.}}
\end{table}

\subsection{NFA experiments}
\begin{figure}[H]
  \centering
  \includegraphics[width=\textwidth]{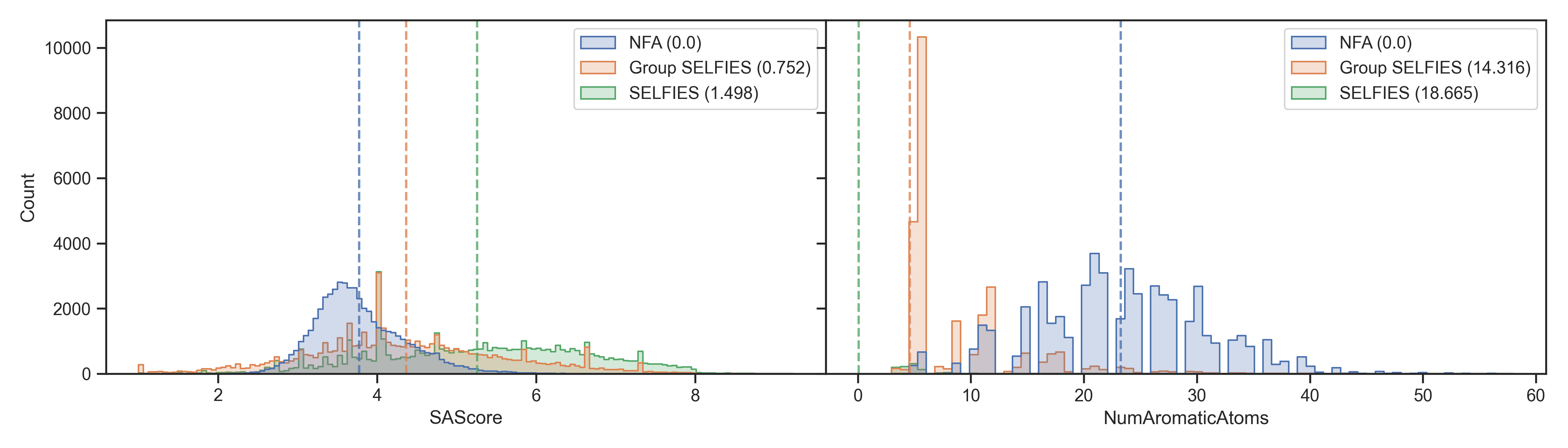}
  \caption{Generated Group SELFIES and regular SELFIES binned by SAScore and number of aromatic atoms. Molecules were generated from the NFA dataset.}
  \label{fig:generated_nfa}
\end{figure}
\label{nfa}
We repeat Experiment \ref{primitive_gen} but for a dataset of nonfullerene acceptors (NFA) \cite{lopez2017design}. This dataset contains many conjugated aromatic systems. In this case, we histogram by SAScore and the number of aromatic atoms, as shown in Figure \ref{fig:generated_nfa}. Generated SELFIES are rarely able to preserve the aromatic systems found in NFA, but generated Group SELFIES preserve much more of the aromatic systems, and can produce molecules with lower and higher SAScore. When binning by number of aromatic atoms, molecules with 0 aromatic atoms are omitted for clarity. About 49\% / 98\% of molecules generated via Group SELFIES / SELFIES have 0 aromatic atoms. Group SELFIES uses a group set containing 74 groups obtained via naïve fragmentation. $N=$ 51281 molecules are generated via Group SELFIES / SELFIES, equal to the size of the NFA dataset.

\subsection{Substring SELFIES}
\label{substrings}

\begin{figure}[H]
  \centering
  \includegraphics[width=\textwidth]{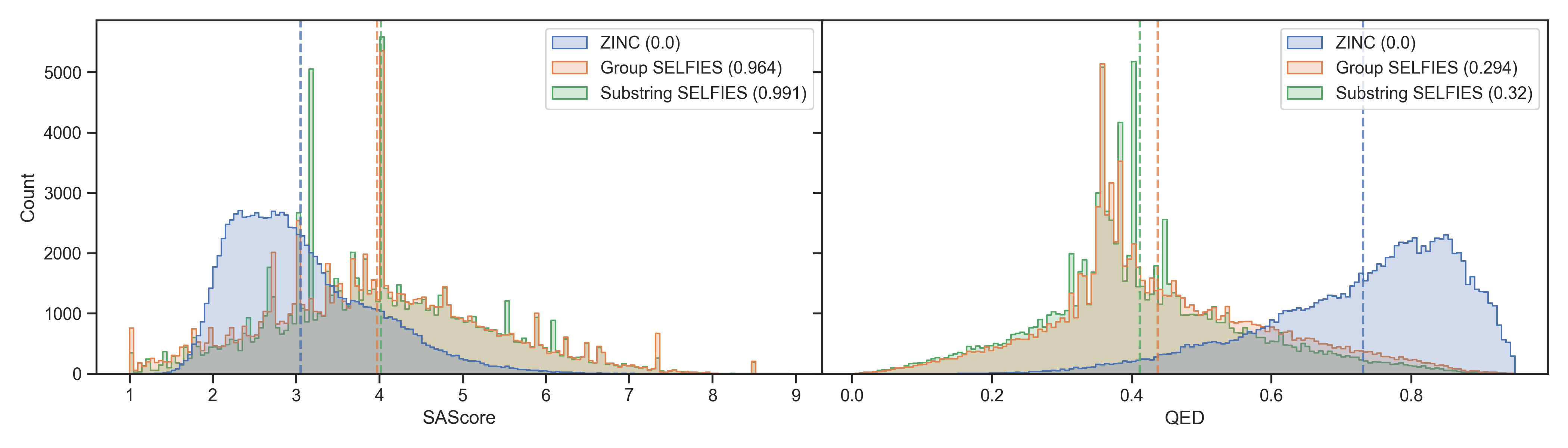}  \caption{Generated Group SELFIES and Substring SELFIES binned by SAScore and QED. Molecules were generated from ZINC-250k.}
  \label{fig:gselfi_subselfi}
\end{figure}

Substring SELFIES is generated by taking a Group SELFIES string and replacing every group token with a SELFIES substring corresponding to it. We compare Substring SELFIES and Group SELFIES in the context of Experiment \ref{primitive_gen} in Figure \ref{fig:gselfi_subselfi} and find that generated Substring SELFIES are on par with generated Group SELFIES.

\subsection{No-Group SELFIES}
\label{no_group}

\begin{figure}[H]
  \centering
  \includegraphics[width=\textwidth]{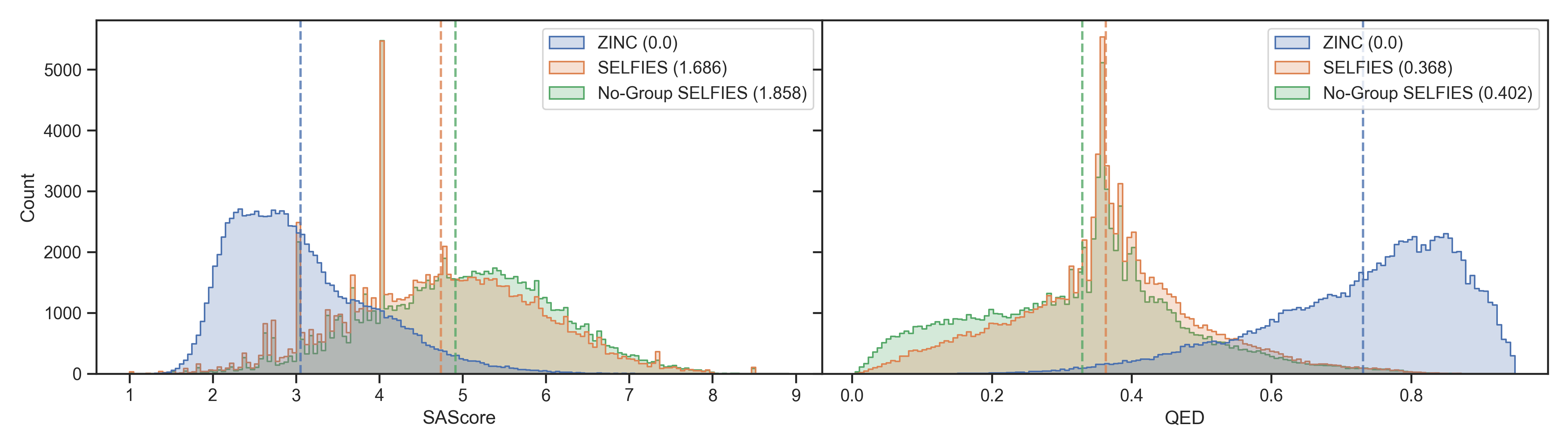}
  \caption{Generated regular SELFIES and No-Group SELFIES binned by SAScore and QED. Molecules were generated from ZINC-250k.}
  \label{fig:selfi_nogselfi}
\end{figure}

No-Group SELFIES is generated by using the Group SELFIES encoder with an empty group set. We compare regular SELFIES to No-Group SELFIES to determine the effect of small changes in the base encoding process, such as branch tokens. In SELFIES, all \texttt{[BranchX]} expect a number token after to determine the length of the branch, whereas in Group SELFIES, all \texttt{[Branch]} tokens do not need a number token, instead using a \texttt{[pop]} token to exit branches. In Figure \ref{fig:selfi_nogselfi}, generated No-Group SELFIES are on par with or worse than generated SELFIES, indicating that the modified encoding process does not contribute much to the overall performance of Group SELFIES.

\subsection{Timing}
\label{timing}

\begin{table}[h]
\centering
\begin{tabular}{ccc}
\toprule
Representation & Encode & Decode \\
\midrule
SELFIES & 0.199 ms & 0.133 ms \\
Group SELFIES & 12.860 ms & 2.494 ms \\
\bottomrule
\\
\end{tabular}
\label{table:timing}

{\center 
\caption{Running time of encoder/decoder for SELFIES and Group SELFIES. The SELFIES encoder is 65x faster than the Group SELFIES encoder, and the SELFIES decoder is 19x faster than the Group SELFIES decoder. Timing is averaged per molecule over the ZINC-250k dataset, using the same group set of 53 groups as the length histogram experiments. Computations were done on a 2021 Apple MacBook Pro with a M1 Pro chip and 32 GB RAM.}}
\end{table}

\end{document}